\documentclass[final,3p,times,authoryear]{elsarticle}

\usepackage{amssymb}
\usepackage{amsmath}

\usepackage{url}
\usepackage{graphicx}
\usepackage[misc]{ifsym}
\usepackage{tcolorbox}
\usepackage{bm}
\usepackage[linesnumbered,ruled,vlined]{algorithm2e}
\usepackage{subcaption}
\SetKwInput{KwInput}{Input}                
\SetKwInput{KwOutput}{Output}              
\SetKwInput{KwParameter}{Parameter}        
\usepackage{array}
\usepackage{rotating}
\usepackage{booktabs}
\usepackage{multirow}
\newcommand{\specialcell}[2][c]{\begin{tabular}[#1]{@{}c@{}}#2\end{tabular}}

\usepackage{enumitem}
\usepackage{hyperref}

\journal{Computer Speech \& Language}

\myfooter[L]{Preprint published by Computer Speech \& Language 
\href{https://doi.org/10.1016/j.csl.2025.101785}{\normalfont\bfseries\ttfamily [https://doi.org/10.1016/j.csl.2025.101785]}}
\begin{document}

\begin{frontmatter}



\title{GenCeption: Evaluate Vision LLMs with Unlabeled Unimodal Data} 


\affiliation[label1]{organization={Microsoft Gaming (ABK)},
            city={Stockholm},
            country={Sweden}}
\affiliation[label2]{organization={EQT Group (Motherbrain)},
            city={Stockholm},
            country={Sweden}}
\affiliation[label3]{organization={Chapter Two},
            city={Stockholm},
            country={Sweden}}
\affiliation[label4]{organization={KTH Royal Institute of Technology},
            city={Stockholm},
            country={Sweden}}
\affiliation[label5]{organization={Télécom Paris},
            city={Palaiseau},
            country={France}}
\affiliation[label6]{organization={Fever Energy},
            city={Stockholm},
            country={Sweden}}
\cortext[cor1]{Equal contribution. Corresponding author: Lele Cao (lelecao@microsoft.com)\\
Source code and leaderboard: \url{https://github.com/llcresearch/GenCeption}\\
This work was initiated during the author's tenure at EQT Motherbrain, with significant parts completed independently thereafter. It represents personal research conducted outside the scope of employment responsibilities. Relevant employers have been informed and have provided consent for publication.}

\author{Lele Cao\fnref{label1,label2}\corref{cor1}}
\author{Valentin Buchner\fnref{label2,label3}\corref{cor1}}
\author{Zineb Senane\fnref{label2,label4,label5,label6}}
\author{Fangkai Yang\fnref{label4}}


\begin{abstract}
Multimodal Large Language Models (MLLMs) are typically assessed using expensive annotated multimodal benchmarks, which often lag behind the rapidly evolving demands of MLLM evaluation. This paper outlines and validates GenCeption, a novel, annotation-free evaluation method that requires only unimodal data to measure inter-modality semantic coherence and inversely assesses MLLMs' tendency to hallucinate. This approach eliminates the need for costly data annotation, minimizes the risk of training data contamination, is expected to result in slower benchmark saturation, and avoids the illusion of emerging abilities. Inspired by the DrawCeption game, GenCeption begins with a non-textual sample and proceeds through iterative description and generation steps. The semantic drift across iterations is quantified using the GC@$T$ metric. While GenCeption is principally applicable to MLLMs across various modalities, this paper focuses on its implementation and validation for Vision LLMs (VLLMs). Based on the GenCeption method, we establish the MMECeption benchmark for evaluating VLLMs, and compare the performance of several popular VLLMs and human annotators. Our empirical results validate GenCeption's effectiveness, demonstrating strong correlations with established VLLM benchmarks. VLLMs still significantly lag behind human performance and struggle especially with text-intensive tasks.
\end{abstract}



\begin{keyword}


multimodal large language model \sep evaluation \sep benchmark
\end{keyword}

\end{frontmatter}



\section{Introduction}
\label{sec:introduction}
Large Language Models (LLMs) demonstrate exceptional abilities in natural language understanding, reasoning, and problem-solving. 
Multimodal LLMs (MLLMs) enhance these capabilities by incorporating multiple modalities, with the visual modality being predominant and highly commercialized~\citep{achiam2023gpt, liu2023improved, jiang2023lion, ye2023mplug}. 
Building on LLMs, MLLMs integrate non-textual modalities, enabling richer interactions and broader applications in real-world scenarios.
However, there is a lack of comprehensive evaluation methods to compare different MLLM architectures and training approaches \citep{fu2023mme}.

In response, the community has developed several MLLM benchmarks, as detailed by~\citet{xu2022multiinstruct,dai2023instructblip,wang2023visionllm,ye2023mplug,li2023evaluating,zhao2023evaluating}. They primarily focus on the visual (i.e., image) and textual input modality due to that VLLMs (Vision LLMs)\footnote{Vision Large Language Models (VLLMs) are a specialized subclass of Multimodal Large Language Models (MLLMs) designed to integrate visual and textual modalities for tasks such as image captioning, visual question answering, and multimodal reasoning. While VLLMs are generally capable of processing various visual data types, their most common input is images, owing to the abundance of annotated image-text datasets and the maturity of image processing technologies.} are the most widely used and readily available MLLMs on the market.
However, these benchmarks face common challenges:
\begin{enumerate}[label=(\arabic*), topsep=0pt, partopsep=0pt, itemsep=.1em]
\item They predominantly rely on multimodal datasets that demand high-quality annotations, which is costly and restrictive in capturing the evolving capabilities of MLLMs~\citep{fu2023mme}. This has been shown to result in increasing speed in benchmark saturation while contemporary models still struggle on trivial real-world tasks~\citep{kiela2021dynabench}. Emerging methods like CrossCheckGPT~\citep{sun2024crosscheckgpt}, designed specifically for MLLM evaluation via cross-system consistency, provide a more relevant, annotation-free alternative. On a broader scope, methods like PRD~\citep{li2023prd} focus on LLM evaluation through peer-based rankings and may be further adapted for MLLM evaluation tasks.
\item MLLM evaluation benchmarks that rely on discrete metrics like accuracy may falsely suggest emergent abilities and do not allow predictable projections of performance improvements from model scaling~\citep{schaeffer2023emergent}. 
\item The evaluation scores may not reflect true performance on real-world tasks due to potential contamination of MLLM training data by benchmark datasets, as reported for LLM pretraining corpora \citep{dodge2021documenting, yang2023rethinking}.
\item The content of one modality is often not needed to answer benchmark questions, as the answer can often be inferred from the question or the MLLM's pretraining knowledge. 
\end{enumerate}
As a consequence of both (3) and (4), some MLLMs can excel on vision QA benchmarks without even being provided the image that is associated with the question.
Existing solutions either only tackle a subset of these challenges, or focus on specific tasks such as image captioning~\citep{lee-etal-2024-fleur}.

\begin{figure*}[]
    \centering
    \includegraphics[width=0.9\textwidth]{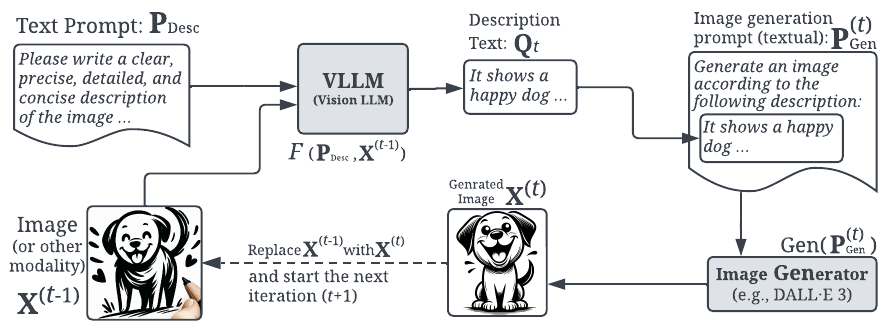}
    \caption{An illustration of the $t$-th iteration in the GenCeption evaluation procedure for VLLMs. Using the image modality as an example, the process begins with an existing image $\mathbf{X}^{(0)}$ sourced from a unimodal image dataset for the first iteration ($t$=1). The VLLM provides a detailed description of the image, which is then used by an image generator to produce $\mathbf{X}^{(t)}$.}
    \label{fig:procedure}
\end{figure*}

We propose GenCeption to address all highlighted challenges involved in the evaluation of task-agnostic MLLMs. GenCeption is designed to be a general evaluation framework that can be applied across modalities. To validate its effectiveness, this paper focuses on Vision LLMs (VLLMs), leveraging the visual modality as an illustrative example. GenCeption addresses challenge (1) by relying on easily accessible unimodal datasets, which allows for cost-effective and scalable benchmark creation. Relying on unimodal datasets additionally addresses challenge (3) and (4), as it allows to easily use previously unseen datasets for MLLM evaluation, and enforces the relevance of the provided modality for excelling at this task. To tackle challenge (2), GenCeption uses the continuous GC@$T$ metric, providing a more nuanced evaluation compared to discrete metrics, allowing for better projections of performance improvements and avoiding the mirage of emergent abilities.

On a high and general level, GenCeption assesses MLLMs' ability to consistently maintain semantic coherence across modalities by iteratively generating and describing non-textual samples and measures the continuous GC@$T$ metric.
This approach simultaneously evaluates the MLLM's tendency to hallucinate, as this inversely correlates with semantic coherence. Further, an MLLM's ability to provide complete yet concise descriptions of non-textual samples measures a diverse range of specialized abilities. For instance, to perform well at describing an image using a limited number of tokens, it is advantageous to be able to reason over people's emotions and intentions behind their actions, infer the current and preceding weather, count objects, and recognize artistic styles. This list can be extended to various abilities depending on the non-textual modality and the content of the samples used during the GenCeption process. The main contributions of this paper are the following:
\begin{itemize}
    \item Proposing GenCeption, an evaluation method that principally allows for using unlabeled unimodal datasets for MLLM evaluation.
    \item Introducing \textit{MMECeption}, a Vision LLM (VLLM) evaluation benchmark utilizing the GenCeption method. \textit{MMECeption} uses the images from the MME benchmark~\citep{fu2023mme}, but without their annotated question and answer pairs.
    \item Evaluating seven leading VLLMs on the \textit{MMECeption} benchmark and comparing results with other popular VLLM benchmarks and human performance. 
\end{itemize}

We will elaborate on the proposed implementation of the GenCeption method, detail our experimental setup, and discuss our findings.

\section{GenCeption}
Our approach, GenCeption, is inspired by a multi-player game DrawCeption\footnote{\url{https://wikipedia.org/wiki/drawception}} (a.k.a.,~Scrawl or Whispernary). 
In this game, the first player is given an image and describes it verbally to the next player. This player then attempts to recreate the image based on the description. The cycle repeats, often resulting in amusing deviations from the original image. The challenge and objective are to maintain the initial information through iterative transitions between verbal descriptions and drawings. Similarly, a proficient Multimodal Language Model (MLLM), which models multiple modalities such as text and images, should excel at this task. Recognizing that MLLMs can encompass modalities beyond just visual cues, such as audio and graphs, we name our approach GenCeption, covering a broader scope than the visually-centric DrawCeption.
For the sake of clarity and alignment with our experiments, we will focus on VLLMs in the remainder of this section to walk through the GenCeption approach. 

While it may not be possible to preserve the initial information perfectly due to varying levels of richness, accuracy, and ambiguity in different modalities, a more capable MLLM will minimize the semantic drift from the original input. 
This contrasts with common benchmarks that aim for complete saturation, highlighting a key advantage of the GenCeption framework: the creation of benchmarks that are more challenging to saturate. With complex initial samples, such as images of real-world scenes or graphs with numerous nodes and edges, this may even result in impossible-to-saturate benchmarks.
Aiming for minimum rather than no semantic drift, this would allow to rank MLLMs relative to each other while continuously leaving space for more performant models.

\subsection{Procedure}
\label{sec:procedure}
Unlike existing MLLM benchmarks (often focused on VLLMs) that rely on multimodal samples, GenCeption is designed to operate on unimodal datasets, significantly streamlining dataset acquisition efforts. For illustrative purposes, we employ the image modality as a representative non-textual modality throughout this exposition. Let us consider an image dataset $\bm{\mathcal{D}}$ comprising images $\mathbf{X}_1, \mathbf{X}_2, \ldots, \mathbf{X}_N$, similar to well-established datasets like ImageNet~\citep{deng2009imagenet}, CIFAR~\citep{krizhevsky2009learning}, and STL~\citep{coates2011analysis}. Without loss of generality, any image from $\bm{\mathcal{D}}$ is denoted as $\mathbf{X}$.

\begin{table}[]
\small\it
\begin{tcolorbox}
Please write a clear, precise, detailed, and concise description of all elements in the image. Focus on accurately depicting various aspects, \textcolor{blue}{including but not limited to the colors, shapes, positions, styles, texts and the relationships between different objects and subjects} in the image. Your description \textcolor{red}{should be thorough enough to guide a professional in recreating this image solely based on your textual representation}. Remember, only include descriptive texts that directly pertain to the contents of the image. You must complete the description using less than 500 words.
\end{tcolorbox}
\caption{The fixed textual prompt $\mathbf{P}_\text{Desc}$ instructs the MLLM to produce a description of the input $\mathbf{X}^{(t-1)}$.}
\label{tab:text-prompt-desc}
\end{table}

GenCeption operates iteratively, from $t=1$ to a pre-defined maximum iteration $t=T$. Each iteration, as depicted in Figure~\ref{fig:procedure}, begins with an image $\mathbf{X}^{(t-1)}$ and yields a new image $\mathbf{X}^{(t)}$. The first iteration ($t=1$) starts with the original image $\mathbf{X}^{(0)}$ from $\bm{\mathcal{D}}$. During any given iteration $t$, the VLLM receives a textual prompt $\mathbf{P}_\text{Desc}$ (Table~\ref{tab:text-prompt-desc}), instructing the VLLM to generate a comprehensive description $\mathbf{Q}_t$ for the input image $\mathbf{X}^{(t-1)}$:
\begin{equation}
\label{eq:mllm}
\mathbf{Q}_t := F(\mathbf{P}_\text{Desc}, \mathbf{X}^{(t-1)}), \text{where} \, F\, \text{denotes the generation function of any VLLMs.}
\end{equation}
Following this, an image generation prompt $\mathbf{P}_{\text{Gen}}^{(t)}$ is constructed as ``{\it\small Generate an image that fully and precisely reflects this description}: $<\mathbf{Q}_t>$''. This prompt guides a pretrained image generation model, such as DALL·E~\citep{ramesh2021zero} or Imagen~\citep{deepmind2023imagegen2}, to create a new image, $\mathbf{X}^{(t)}$:
\begin{equation}
\label{eq:gen}
\mathbf{X}^{(t)} := \text{Gen}(\mathbf{P}_\text{Gen}^{(t)}),
\end{equation}
where Gen(·) signifies the chosen image generator. Each subsequent iteration $t+1$ starts with the image $\mathbf{X}^{(t)}$ generated in the previous iteration. Upon completing all iterations, we obtain a series of $T+1$ images: $\mathbf{X}^{(0)}, \mathbf{X}^{(1)}, \ldots, \mathbf{X}^{(T)}$, with the initial image being the original and the rest sequentially produced across the iterations.

The textual prompt $\mathbf{P}_\text{Desc}$ is intentionally kept short and concise to minimize potential variations in model behaviours due to susceptibility to prompt composition~\citep{loya2023exploring}.

\subsection{Metric: GC@$T$}
\label{sec:metric}
Our primary objective is to measure the semantic divergence of each generated image $\mathbf{X}^{(t)}$ (for $t \in \{1, 2, \ldots, T\}$) from the original image $\mathbf{X}^{(0)}$.
We use a pretrained image encoder, such as ViT~\citep{dosovitskiy2021an}, to transform all images, resulting in $T+1$ image embeddings denoted as $\mathbf{z}^{(0)}, \mathbf{z}^{(1)}, \ldots, \mathbf{z}^{(T)}$, where $\mathbf{z}^{(t)} := \text{Enc}(\mathbf{X}^{(t)})$.
We then compute the cosine similarity between $\mathbf{z}^{(0)}$ and each $\mathbf{z}^{(t)}$ (for $t \in \{1, 2, \ldots, T\}$), yielding $T$ similarity scores: $s^{(1)}, s^{(2)}, \ldots, s^{(T)}$. Here, $s^{(t)}\in[-1.0, 1.0]$ approximates the level of semantic drift in the $t$-th iteration of the GenCeption procedure.
To quantify the overall speed and magnitude of semantic drift, we calculate the GenCeption score over $T$ iterations, denoted as GC@$T \in [-1.0, 1.0]$, as follows:
\begin{equation}
\label{eq:gct}
\text{GC@}T := \frac{\sum_{t=1}^{T}(t\cdot s^{(t)})}{\sum_{t=1}^{T}t}.
\end{equation}
This is a normalized and continuous metric that weights later iterations more heavily for two reasons: (1) similar to the DrawCeption game, the last image's deviation from the initial image is most telling; (2) we aim to capture performance and dynamics across the entire iterative sequence.
A high GC@$T$ value signifies an exceptional ability to maintain inter-modal (text-image) semantic congruence, effectively curbing the propensity for rapid or extensive deviation from the semantics encapsulated in the original image.
Notably, GC@$1$ is equivalent to $s^{(1)}$.
For the pseudo code detailing the GenCeption procedure and the calculation of the average GC@$T$ metric over the entire dataset $\bm{\mathcal{D}}$, see Algorithm~\ref{algo:genception}.

For the special case of VLLMs that are evaluated in this study, we additionally replace using ViT embeddings and cosine similarity with the Frechet Inception Distance (FID), a metric commonly used to evaluate image generation models~\citep{heusel2017gans}. The FID is calculated between the original dataset of images $\bm{\mathcal{D}}^{(0)}$, and the images generated from the respective dataset using the GenCeption process $\bm{\mathcal{D}}^{(t)}$, yielding $T$ FID scores: $\text{fid}^{(1)}, \text{fid}^{(2)}, \ldots, \text{fid}^{(T)}$. The GC$_\text{FID}$@$T$ score is then calculated as: 
\begin{equation}
\label{eq:gctfid}
\text{GC}_\text{FID}\text{@}T := \frac{\sum_{t=1}^{T}(t\cdot \text{fid}^{(t)})}{\sum_{t=1}^{T}t}.
\end{equation}

As the FID indicates a distance rather than a similarity between two sets of images, a lower distance indicates better performance, and consequently a lower GC$_\text{FID}$@$T$ score indicates a more capable VLLM.

\begin{algorithm}[t!]
\SetAlgoLined
\SetNlSty{}{}{:}
\DontPrintSemicolon
{
\KwInput{VLLM to be evaluated, a unimodal dataset \vspace{2pt}$\bm{\mathcal{D}}$: $\small\smash{\mathbf{X}_1^{(0)},\ldots,\mathbf{X}_n^{(0)},\ldots,\mathbf{X}_N^{(0)}}$, fixed textual prompt $\mathbf{P}_\text{Desc}$, a sample generator Gen(·), and a sample encoder Enc(·)}
\KwOutput{Average GC@$T$ metric over $\bm{\mathcal{D}}$}

\KwParameter{The number of iterations $T$}

GC@$T$ = 0

\For{$(n = 1; n \leq N; n++)$}{

    $\mathbf{z}^{(0)}$ := Enc($\mathbf{X}_n^{(0)}$);

    \For{$(t = 1; t \leq T; t++)$}{
    
        Generate description $\mathbf{Q}_t$ for $\mathbf{X}_n^{(t-1)}$ using \eqref{eq:mllm};
    
        Create $\small\smash{\mathbf{P}_{\text{Gen}}^{(t)}}$ using $\mathbf{Q}_t$;
    
        Generate $\mathbf{X}_n^{(t)}$ according to \eqref{eq:gen};

        $s^{(t)}$ := CosineSimilarity($\small\smash{\mathbf{z}^{(0)}}$, Enc($\small\smash{\mathbf{X}_n^{(t)}}$));
        
    }

    Calculate GC@$T$ += $\small\smash{\sum_{t=1}^{T}(t\cdot s^{(t)})/\sum_{t=1}^{T}t}$; \eqref{eq:gct}
}

\textbf{return} GC@$T$ / N;
\caption{Calculate GC@$T$ via GenCeption for a specific VLLM under evaluation}
\label{algo:genception}
}
\end{algorithm}

\begin{sidewaystable*}
\centering
\footnotesize
\addtolength{\tabcolsep}{-2pt}
\renewcommand{\arraystretch}{1.2}
\begin{tabular}{c c |rrr|rrr|rrr|rrr|rrr|rrr|rrr}
\toprule
\multicolumn{2}{c|}{\multirow{3.5}{*}{\specialcell{Sample group\\ \& category}}}&\multicolumn{3}{c|}{Gemini1.5-Pro}&\multicolumn{3}{c|}{Claude3-Opus}&\multicolumn{3}{c|}{GPT-4o}&\multicolumn{3}{c|}{GPT-4V}&\multicolumn{3}{c|}{mPLUG-Owl2}&\multicolumn{3}{c|}{LLaVA-13B}&\multicolumn{3}{c}{LLaVA-7B}\\
\cmidrule{3-23}
&&\specialcell{{\scriptsize MME$\uparrow$}\\ 
{\scriptsize $\sum$ACC}}&\specialcell{{\scriptsize GC\!@\!3$\uparrow$}\\ {\scriptsize Cosine}}&\specialcell{{\scriptsize GC\!@\!3$\downarrow$}\\ {\scriptsize FID}}&\specialcell{{\scriptsize MME$\uparrow$}\\ 
{\scriptsize $\sum$ACC}}&\specialcell{{\scriptsize GC\!@\!3$\uparrow$}\\ {\scriptsize Cosine}}&\specialcell{{\scriptsize GC\!@\!3$\downarrow$}\\ {\scriptsize FID}}&\specialcell{{\scriptsize MME$\uparrow$}\\ 
{\scriptsize $\sum$ACC}}&\specialcell{{\scriptsize GC\!@\!3$\uparrow$}\\ {\scriptsize Cosine}}&\specialcell{{\scriptsize GC\!@\!3$\downarrow$}\\ {\scriptsize FID}}&\specialcell{{\scriptsize MME$\uparrow$}\\ {\scriptsize $\sum$ACC}}&\specialcell{{\scriptsize GC\!@\!3$\uparrow$}\\ {\scriptsize Cosine}}&\specialcell{{\scriptsize GC\!@\!3$\downarrow$}\\ {\scriptsize FID}}&\specialcell{{\scriptsize MME$\uparrow$}\\ {\scriptsize $\sum$ACC}}&\specialcell{{\scriptsize GC\!@\!3$\uparrow$}\\ {\scriptsize Cosine}}&\specialcell{{\scriptsize GC\!@\!3$\downarrow$}\\ {\scriptsize FID}}&\specialcell{{\scriptsize MME$\uparrow$}\\ {\scriptsize $\sum$ACC}}&\specialcell{{\scriptsize GC\!@\!3$\uparrow$}\\ {\scriptsize Cosine}}&\specialcell{{\scriptsize GC\!@\!3$\downarrow$}\\ {\scriptsize FID}}&\specialcell{{\scriptsize MME$\uparrow$}\\ {\scriptsize $\sum$ACC}}&\specialcell{{\scriptsize GC\!@\!3$\uparrow$}\\ {\scriptsize Cosine}}&\specialcell{{\scriptsize GC\!@\!3$\downarrow$}\\ {\scriptsize FID}} \\
\midrule 

\parbox[t]{3.5mm}{\multirow{13}{*}{\rotatebox[origin=c]{270}{visual-intensive samples}}} &
\multicolumn{1}{|c|}{Existence}     &190.0&{\bf 0.437}&269.8&183.3&0.382&273.7&{\bf 195.0}&0.400&266.5&175.0&0.422&{\bf 265.1}&185.0&0.323&296.9&{\bf 195.0}&0.305&322.0&{\bf 195.0}&0.308&318.1\\
    &\multicolumn{1}{|c|}{Count}    &148.3&0.389&{\bf 272.6}&116.7&0.348&285.3&{\bf 190.0}&0.388&277.5&153.3&\textbf{0.404}&277.4&160.0&0.299&316.1&165.0&0.294&319.7&148.3&0.253&326.1\\
    &\multicolumn{1}{|c|}{Position} &105.0&0.357&253.7&76.7&0.357&266.1&{\bf 145.0}&0.398&260.6&85.0&\textbf{0.408}&{\bf 253.3}&75.0&0.306&294.0&135.0&0.255&298.9&123.3&0.285&286.0\\
    &\multicolumn{1}{|c|}{Color}    &175.0&{\bf 0.474}&{\bf 234.8}&118.3&0.385&267.6&{\bf 180.0}&0.421&246.1&141.7&0.403&243.7&138.3&0.290&310.1&165.0&0.300&305.6&170.0&0.284&304.5\\
    &\multicolumn{1}{|c|}{Poster}   &175.2&{\bf 0.374}&206.0&149.7&0.360&206.0&{\bf 192.2}&0.335&{\bf 203.7}&187.8&0.324&209.4&154.8&0.243&209.2&163.6&0.215&240.9&154.1&0.214&244.3\\
    &\multicolumn{1}{|c|}{Celebrity} &{\bf 169.4}&{\bf 0.362}&191.0&77.6&0.317&192.5&46.8&0.331&193.3&53.5&0.332&{\bf 189.1}&167.9&0.232&211.3&144.4&0.206&223.7&153.2&0.188&233.6\\
    &\multicolumn{1}{|c|}{Scene}    &147.0&{\bf 0.423}&{\bf 173.7}&149.8&0.374&174.7&148.5&0.401&{\bf 171.5}&141.2&0.393&173.7&157.8&0.299&194.4&{\bf 162.8}&0.277&198.0&160.8&0.266&196.3\\
    &\multicolumn{1}{|c|}{Landmark} &{\bf 176.8}&{\bf 0.375}&182.1&113.0&0.344&188.9&175.5&0.372&{\bf 182.0}&104.0&0.353&182.6&158.8&0.275&206.0&150.8&0.242&224.3&154.8&0.252&214.4\\
    &\multicolumn{1}{|c|}{Artwork}  &{\bf 152.2}&0.412&171.1&136.8&0.385&170.4&144.0&0.415&{\bf 169.2}&115.0&\textbf{0.421}&170.3&136.0&0.252&202.2&98.8&0.212&213.3&110.0&0.210&215.3\\
    &\multicolumn{1}{|c|}{Commonsense}    &150.0&0.464&216.5&115.0&0.432&210.4&{\bf 174.3}&0.448&213.9&155.0&\textbf{0.471}&{\bf 208.1}&127.9&0.353&237.2&115.7&0.334&248.9&117.1&0.294&254.4\\
    \cmidrule{2-23}
    &\multicolumn{1}{|c|}{Visual Mean}&158.9&{\bf 0.407}&{\bf 217.1}&123.7&0.368&223.6&{\bf 159.1}&0.391&218.4&131.2&0.393&217.3&146.2&0.287&247.7&149.7&0.264&259.5&148.7&0.255&259.3\\
    &\multicolumn{1}{|c|}{Visual Rank}&2&{\bf 1}&1&7&4&4&{\bf 1}&3&3&6&2&2&5&5&5&3&6&7&5&7&6\\
\midrule
\parbox[t]{2mm}{\multirow{6}{*}{\rotatebox[origin=c]{270}{{\footnotesize textual-intensive}}}} &
    \multicolumn{1}{|c|}{Code reasoning}       &117.5&0.213&310.0&70.0&0.245&{\bf 267.4}&{\bf 182.5}&{\bf 0.255}&299.7&147.5&0.193&302.9&65.0&0.176 &327.6&55.0&0.144&323.5&50.0&0.107&398.2\\
    &\multicolumn{1}{|c|}{Numerical calc.}       &110.0&0.268&346.5&67.5&0.229&349.3&{\bf 170.0}&{\bf 0.282}&346.4&80.0&0.240&{\bf 322.5}&45.0&0.192 &362.0&35.0&0.195&367.4&50.0&0.155&366.0\\
    &\multicolumn{1}{|c|}{Text translation}&162.5&{\bf 0.240}&334.6&45.0&0.236&362.5&{\bf 192.5}&0.211&{\bf 326.9}&55.0&0.157&368.0&112.5&0.081 &365.2&85.0&0.116&352.3&65.0&0.111&424.4\\
    &\multicolumn{1}{|c|}{OCR}        &170.0&0.367&{\bf 233.2}&167.5&0.362&245.5&{\bf 192.5}&0.362&246.2&177.5&\textbf{0.393}&238.0&102.5&0.276 &255.4&95.0&0.239&270.6&65.0&0.222&283.7\\
    \cmidrule{2-23}
    &\multicolumn{1}{|c|}{Textual Mean}&140.0&0.272&306.1&87.5&0.268&306.2&{\bf 184.4}&{\bf 0.278}&{\bf 304.8}&115.0&0.246&307.9&81.3&0.181&327.6&67.5&0.174&328.5&57.5&0.149&368.1\\
    &\multicolumn{1}{|c|}{Textual Rank}&2&2&2&4&3&3&{\bf 1}&{\bf 1}&{\bf 1}&3&4&4&5&5&5&6&6&6&7&7&7\\
\midrule
\multicolumn{2}{c|}{Overall Mean}
    &153.5&{\bf 0.368}&{\bf 242.5}&113.3&0.340&247.2&{\bf 166.3}&0.359&243.1&126.5&0.351&243.2&127.6&0.257&270.5&126.1&0.238&279.2&122.6&0.225&290.1\\
    \multicolumn{2}{c|}{Overall Rank}
    &2&{\bf 1}&{\bf 1}&7&4&4&{\bf 1}&2&2&4&3&3&3&5&5&5&6&6&6&7&7\\
\midrule
    \multicolumn{2}{c|}{HallusionBench\textsuperscript{*}}
    &\multicolumn{3}{c|}{45.2 (rank=3)}&\multicolumn{3}{c|}{37.8 (rank=4)}&\multicolumn{3}{c|}{\bf 51.7 (rank=1)}&\multicolumn{3}{c|}{46.5 (rank=2)}&\multicolumn{3}{c|}{25.7 (rank=6)}&\multicolumn{3}{c|}{24.5 (rank=7)}&\multicolumn{3}{c}{27.6 (rank=5)}\\
    \multicolumn{2}{c|}{MMStar\textsuperscript{*}}
    &\multicolumn{3}{c|}{38.6 (rank=5)}&\multicolumn{3}{c|}{45.7 (rank=3)}&\multicolumn{3}{c|}{\bf 61.6 (rank=1)}&\multicolumn{3}{c|}{47.7 (rank=2)}&\multicolumn{3}{c|}{34.8 (rank=6)}&\multicolumn{3}{c|}{40.1 (rank=4)}&\multicolumn{3}{c}{34.6 (rank=7)}\\
    \multicolumn{2}{c|}{SEEDBench (Test)\textsuperscript{*}}
    &\multicolumn{3}{c|}{70.7 (rank=3)}&\multicolumn{3}{c|}{64.0 (rank=7)}&\multicolumn{3}{c|}{\bf 76.4 (rank=1)}&\multicolumn{3}{c|}{71.6 (rank=2)}&\multicolumn{3}{c|}{64.5 (rank=6)}&\multicolumn{3}{c|}{67.9 (rank=4)}&\multicolumn{3}{c}{66.4 (rank=5)}\\
    \multicolumn{2}{c|}{AI2D\textsuperscript{*}}
    &\multicolumn{3}{c|}{70.2 (rank=4)}&\multicolumn{3}{c|}{70.6 (rank=3)}&\multicolumn{3}{c|}{\bf 82.2 (rank=1)}&\multicolumn{3}{c|}{75.5 (rank=2)}&\multicolumn{3}{c|}{55.7 (rank=7)}&\multicolumn{3}{c|}{61.3 (rank=5)}&\multicolumn{3}{c}{55.9 (rank=6)}\\
    \multicolumn{2}{c|}{OpenCompass\textsuperscript{*}}
    &\multicolumn{3}{c|}{62.7 (rank=3)}&\multicolumn{3}{c|}{57.7 (rank=4)}&\multicolumn{3}{c|}{\bf 66.3 (rank=1)}&\multicolumn{3}{c|}{63.3 (rank=2)}&\multicolumn{3}{c|}{46.3 (rank=7)}&\multicolumn{3}{c|}{48.8 (rank=5)}&\multicolumn{3}{c}{46.7 (rank=6)}\\
\bottomrule
\end{tabular}%
\begin{flushleft}
\scriptsize

\ * Results are sourced from \url{https://huggingface.co/spaces/opencompass/open_vlm_leaderboard} as of 2024-04-25 (except GPT-4o) and 2024-05-23 (for GPT-4o).

\ $\downarrow$ Results are obtained using FID to measure the similarity between images. \url{https://github.com/GaParmar/clean-fid}
\end{flushleft}
\caption{Evaluation results of GC@$3$, MME, HallusionBench and OpenCompass on visual(Vis)-intensive and textual(Text)-intensive images. Best results per metric and category (over different MLLMs) are \textbf{bolded}.}
\label{table:main-visual-text-intensive}
\end{sidewaystable*}

\section{Experiments}
\label{sec:experiments}

We run several extensive experiments to validate the GenCeption method by comparing the GC@$T$ scores achieved by several VLLMs to the scores they achieve on carefully crafted established benchmarks and to average human performance. Although GenCeption's design merely requires unimodal image datasets, we employ the same data as used by a recent and well-validated MLLM benchmark, MME~\citep{fu2023mme}. While we discard the annotated question-answer pairs associated with the images in this benchmark, this provides us with the ability (1) to facilitate direct comparison with metrics that include textual QA (question-answering) annotations, and (2) to enable a detailed assessment of MLLM performance across MME's 14 meticulously crafted sample categories. Attributing this newly created benchmark to the MME dataset and the GenCeption method, we refer to it as the \textit{MMECeption} benchmark.

We select seven VLLMs -- Gemini1.5-Pro~\citep{reid2024gemini}, GPT-4o~\citep{openai2024gpt4o} , GPT-4V~\citep{achiam2023gpt}, Claude3-Opus~\citep{anthropic2023claude}, LLaVA-7B/13B~\citep{liu2023improved} and mPLUG-Owl2~\citep{ye2023mplug} -- based on their superior performance on the OpenCompass multimodal leaderboard~\citep{2023opencompass}, which incorporates a comprehensive set of benchmarks like MME~\citep{fu2023mme},  HallusionBench~\citep{liu2023hallusionbench}, MMStar~\citep{chen2024we}, SeedBench~\citep{li2023seed}, and AI2D~\citep{kembhavi2016diagram}.
We use DALL$\cdot$E as the default image generation model. To prevent potential bias towards OpenAI-developed VLLMs, which might have had access to DALL$\cdot$E-generated images during their training, we perform an additional evaluation of all VLLMs on the GC@$1$ score using Imagen2 as an image generation model.
We set the temperature parameter to 0 in both the VLLMs and image generators to minimize the stochasticity in model outputs.

As humans are well versed at integrating vision and language modalities, we aim to quantify average human performance on the \textit{MMECeption} benchmark. As the GenCeption procedure is a labor-intensive and time-consuming task for humans, we randomly select 5 images from each MME category, and by providing human annotators with the same prompts as defined in Table~\ref{tab:text-prompt-desc}, collect results and calculate the GC@$1$ metric. 
Five human annotators (3 master students, 1 lecturer, and 1 artist) were recruited to describe one image of each category such that each image in a category is described by a different person to mitigate personal performance differences. The annotators were given 14 weeks to perform this task and were awarded a generous reimbursement of €40 each to ensure sufficient dedication. All annotators were either native English speakers or fluent at a professional level.

\subsection{Quantitative results}
\label{sec:quantitative-results}
We partition the 14 MME categories into two groups based on content type: visual-intensive (10 categories: {\it existence, count, position, color, poster, celebrity, scene, landmark, artwork,} and {\it commonsense reasoning}) and textual-intensive (4 categories: {\it code reasoning, numerical calculation, text translation,} and {\it OCR}).
GC@$3$ scores on the \textit{MMECeption} benchmark and accuracy on the original MME benchmark are reported per category and as aggregations in Table~\ref{table:main-visual-text-intensive}.
Additionally, we include the scores and ranks of all evaluated VLLMs on the OpenCompass~\citep{2023opencompass}, MME~\citep{fu2023mme},  HallusionBench~\citep{liu2023hallusionbench}, MMStar~\citep{chen2024we}, SeedBench~\citep{li2023seed}, and AI2D~\citep{kembhavi2016diagram} leaderboards.
Notably, Gemini1.5-Pro leads our rankings, followed by GPT-4o, GPT-4v, Claude3-Opus, mPLUG-Owl2, and LLaVA-13B/7B. The GenCeption method shows robustness to the similarity metric used, as the overall ranking remains identical when using cosine similarity or FID distance for calculating GC@$T$ scores. 

\begin{figure}[t!]
    \centering
    \centering
    \includegraphics[width=.5\linewidth]{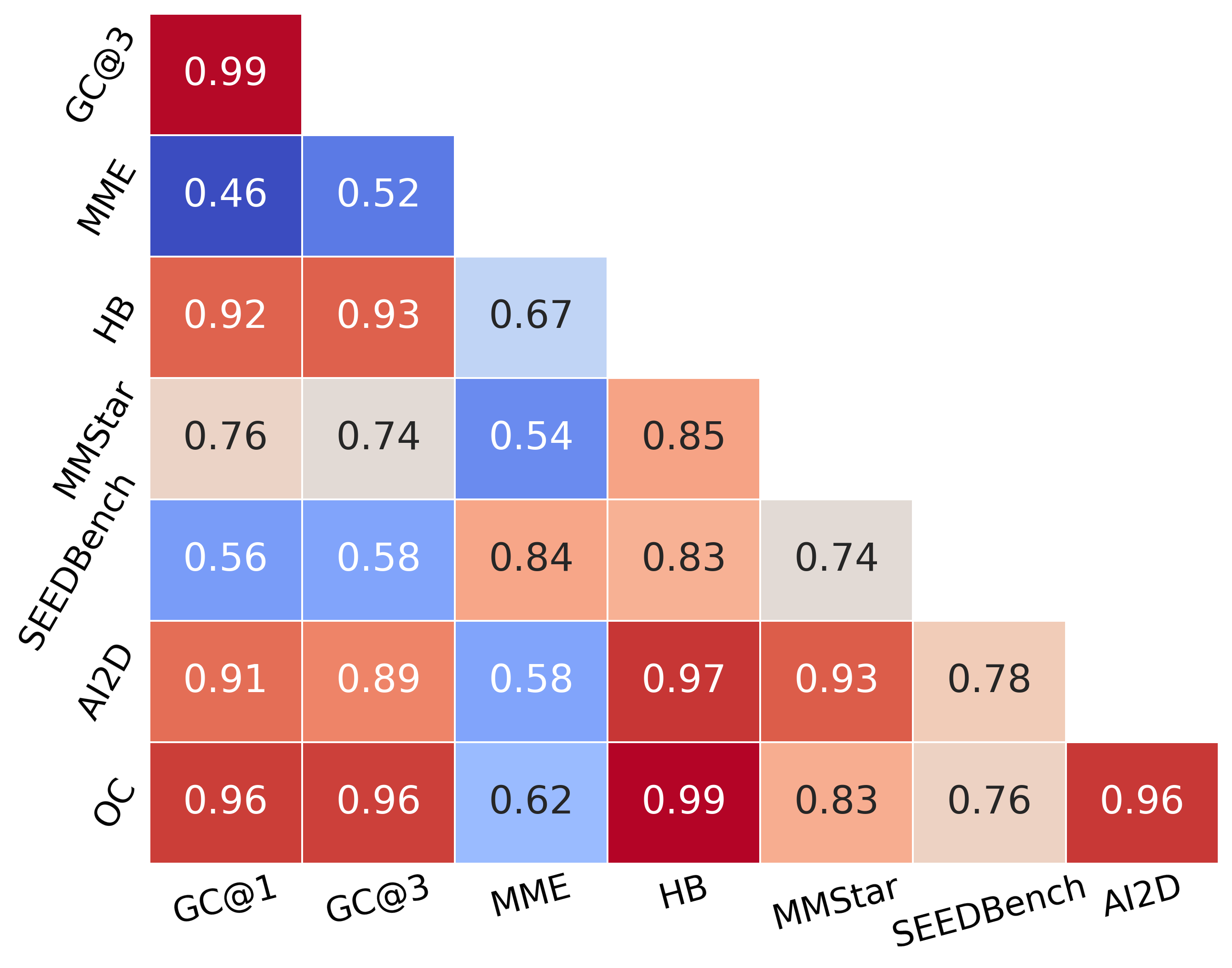}
    \caption{Correlation Matrix of GC@$1$ and GC@$3$ scores on \textit{MMECeption}, and several other benchmarks.}
    \label{fig:correlations}
\end{figure}

\begin{table}[ht]
\centering
\begin{tabular}{l|cccc}
\hline
            & GC@1 & GC@3 & MME & HallusionBench \\ \hline
GC@1 (essentially the same as ~$s^{(1)}$) & -    & 0.99 & 0.46 & 0.92 \\
GC@3        & 0.99 & -    & 0.52 & 0.93 \\
$s^{(3)}$ (GC@3 without weighting by $t$)       & 0.98 & 0.99 & 0.46 & 0.94 \\
CrossCheckGPT (Image-to-text) & 0.96 & 0.94 & 0.42 & 0.97 \\ \hline
\end{tabular}
\caption{Correlation matrix comparing GC@$1$, GC@$3$, $s^{(3)}$ (GC@$3$ without temporal weighting), and CrossCheckGPT with established benchmarks MME and HallusionBench.}
\label{tab:metric_comparison}
\end{table}

\begin{table*}[!t]
\centering
\small
\addtolength{\tabcolsep}{-1.8pt}
\renewcommand{\arraystretch}{0.9}
\begin{tabular}{c c |rr|rr|rr|rr|rr|rr|rr}
\toprule
\multicolumn{2}{c|}{\multirow{2.5}{*}{\specialcell{Sample group\\ \& category}}}&\multicolumn{2}{c|}{\scriptsize Gemini1.5-Pro}&\multicolumn{2}{c|}{\scriptsize Claude3-Opus}&\multicolumn{2}{c|}{\scriptsize GPT-4o}&\multicolumn{2}{c|}{\scriptsize GPT-4V}&\multicolumn{2}{c|}{\scriptsize mPLUG-Owl2}&\multicolumn{2}{c|}{\scriptsize LLaVA-13B}&\multicolumn{2}{c}{\scriptsize LLaVA-7B}\\
\cmidrule{3-16}
&&{\scriptsize Dalle3}&{\scriptsize Imgn2}&
{\scriptsize Dalle3}&{\scriptsize Imgn2}&{\scriptsize Dalle3}&{\scriptsize Imgn2}&{\scriptsize Dalle3}&{\scriptsize Imgn2}&{\scriptsize Dalle3}&{\scriptsize Imgn2}&{\scriptsize Dalle3}&{\scriptsize Imgn2}&{\scriptsize Dalle3}&{\scriptsize Imgn2} \\
\midrule 

\parbox[t]{2mm}{\multirow{13}{*}{\rotatebox[origin=c]{90}{visual-intensive samples}}} &
\multicolumn{1}{|c|}{Existence}     &0.505&0.529&0.500&{\bf 0.532}&{\bf 0.536}&0.521&0.505&0.530&0.427&0.515&0.416&0.485&0.418&0.506\\
    &\multicolumn{1}{|c|}{Count}    &0.456&0.489&0.466&0.490&0.456&0.494&{\bf 0.498}&{\bf 0.506}&0.378&0.463&0.408&0.466&0.341&0.416\\
    &\multicolumn{1}{|c|}{Position} &{\bf 0.511}&{\bf 0.491}&0.495&0.480&0.469&0.460&0.501&0.473&0.346&0.452&0.359&0.454&0.350&0.402\\
    &\multicolumn{1}{|c|}{Color}    &{\bf 0.545}&{\bf 0.525}&0.489&0.501&0.480&0.473&0.506&0.490&0.345&0.471&0.420&0.457&0.318&0.436\\
    &\multicolumn{1}{|c|}{Poster}   &{\bf 0.455}&{\bf 0.388}&0.450&0.381&0.445&0.383&0.444&0.365&0.338&0.357&0.303&0.312&0.305&0.266\\
    &\multicolumn{1}{|c|}{Celebrity} &0.417&0.384&0.424&0.382&0.418&0.373&{\bf 0.433}&{\bf 0.389}&0.319&0.336&0.284&0.317&0.263&0.313\\
    &\multicolumn{1}{|c|}{Scene}    &{\bf 0.511}&{\bf 0.490}&0.504&0.478&0.482&0.474&0.497&0.474&0.385&0.417&0.355&0.404&0.350&0.392\\
    &\multicolumn{1}{|c|}{Landmark} &{\bf 0.500}&0.485&0.460&{\bf 0.492}&0.494&0.479&0.458&0.480&0.363&0.351&0.376&0.357&0.334&0.333\\
    &\multicolumn{1}{|c|}{Artwork}  &0.494&0.454&{\bf 0.508}&{\bf 0.461}&0.500&0.455&0.504&0.455&0.333&0.385&0.308&0.333&0.294&0.304\\
    &\multicolumn{1}{|c|}{Common.}    &0.545&0.531&0.535&0.507&0.562&0.526&{\bf 0.563}&{\bf 0.535}&0.425&0.493&0.429&0.473&0.417&0.458\\
    \cmidrule{2-16}
    &\multicolumn{1}{|c|}{Vis Mean}&{\bf 0.494}&{\bf 0.477}&0.483&0.470&0.484&0.464&0.491&0.470&0.366&0.424&0.366&0.406&0.339&0.383\\
    &\multicolumn{1}{|c|}{Vis Rank}&{\bf 1}&{\bf 1}&4&2&3&4&2&2&5&5&5&6&7&7\\
\midrule
\parbox[t]{2mm}{\multirow{6}{*}{\rotatebox[origin=c]{90}{{\footnotesize textual-intensive}}}} &
    \multicolumn{1}{|c|}{Code}       &0.364&0.177&0.304&0.180&{\bf 0.395}&0.179&0.333&{\bf 0.263}&0.281&0.100 &0.260&0.168&0.186&0.108\\
    &\multicolumn{1}{|c|}{Numerical}   &0.322&0.417&0.333&0.389&{\bf 0.366}&{\bf 0.456}&0.325&0.383&0.322&0.225 &0.336&0.265&0.259&0.222\\
    &\multicolumn{1}{|c|}{Text trans.}&0.396&0.227&0.356&0.258&{\bf 0.444}&{\bf 0.277}&0.359&0.238&0.173&0.052 &0.200&0.118&0.212&0.073\\
    &\multicolumn{1}{|c|}{OCR}        &0.462&{\bf 0.500}&{\bf 0.486}&0.448&0.421&0.441&0.482&0.417&0.358&0.384 &0.368&0.385&0.351&0.320\\
    \cmidrule{2-16}
    &\multicolumn{1}{|c|}{Text Mean}&0.386&0.330&0.370&0.319&{\bf 0.407}&{\bf 0.338}&0.375&0.325&0.284&0.190&0.291&0.234&0.252&0.181\\
    &\multicolumn{1}{|c|}{Text Rank}&2&2&4&4&{\bf 1}&{\bf 1}&3&3&6&6&5&5&7&7\\
\midrule
\multicolumn{2}{c|}{Overall Mean}
    &{\bf 0.463}&{\bf 0.435}&0.451&0.427&0.462&0.428&0.458&0.428&0.343&0.357&0.344&0.357&0.314&0.325\\
    \multicolumn{2}{c|}{Overall Rank}
    &{\bf 1}&{\bf 1}&4&4&2&2&3&2&6&5&5&5&7&7\\
\bottomrule
\end{tabular}%
\caption{The impact of different image encoders, DALL·E~3 (Dalle3) vs. Imagen~2 (Imgn2), on GC@$1$ score. Best results per configuration and category (over different VLLMs) are \textbf{bolded}.}
\label{table:main-gc1-dalle3-imagen2}
\end{table*}

\begin{table*}[!t]
\centering
\small
\addtolength{\tabcolsep}{-0.8pt}
\begin{tabular}{c c |r|r|r|r|r|r|r|r|r}
\toprule
\multicolumn{2}{c|}{\specialcell{Sample group\\ \& category}}&\specialcell{Gemini1.5\\ -Pro}&\specialcell{Claude3\\ -Opus}&GPT-4o&GPT-4V&\specialcell{mPLUG\\ -Owl2}&\specialcell{LLaVA\\ -13B}&\specialcell{LLaVA\\ -7B}&Human&\specialcell{\scriptsize$\Delta$\% between\\ \scriptsize \textbf{human} \& \underline{best}}\\
\midrule 

\parbox[t]{2mm}{\multirow{12}{*}{\rotatebox[origin=c]{90}{visual-intensive samples}}} &
\multicolumn{1}{|c|}{Existence}     &\underline{0.5841}&0.4563&0.5578&0.5434&0.3967&0.3524&0.3782&{\bf 0.6402}&+ 9.6045\%\\
    &\multicolumn{1}{|c|}{Count}    &0.4140&0.3799&0.2725&\underline{0.4882}&0.2364&0.3535&0.2038&{\bf 0.5476}&+ 12.1671\%\\
    &\multicolumn{1}{|c|}{Position} &0.5546&0.4959&0.4086&\underline{0.5639}&0.3527&0.4285&0.3899&{\bf 0.6409}&+ 13.6549\%\\
    &\multicolumn{1}{|c|}{Color}    &\underline{0.7081}&0.6206&0.6139&0.5516&0.4047&0.4314&0.3506&{\bf 0.8380}&+ 18.3449\%\\
    &\multicolumn{1}{|c|}{Poster}   &\underline{0.5046}&0.4362&0.4939&0.4681&0.3998&0.3208&0.2905&{\bf 0.5456}&+ 8.1252\%\\
    &\multicolumn{1}{|c|}{Celebrity} &0.4182&0.3988&0.4369&\underline{0.4447}&0.3714&0.2545&0.2160&{\bf 0.4671}&+ 5.0371\%\\
    &\multicolumn{1}{|c|}{Scene}    &\underline{0.6080}&0.5828&0.5229&0.5919&0.4842&0.3906&0.4057&{\bf 0.6236}&+ 2.5658\%\\
    &\multicolumn{1}{|c|}{Landmark} &0.4903&0.4932&0.5236&\underline{0.5702}&0.3613&0.4174&0.3845&{\bf 0.6045}&+ 6.0154\%\\
    &\multicolumn{1}{|c|}{Artwork}  &0.3725&\underline{0.5304}&0.5297&0.5252&0.2938&0.2924&0.2336&{\bf 0.5421}&+ 2.2059\%\\
    &\multicolumn{1}{|c|}{Commonsense}    &0.4338&\underline{0.5375}&0.5047&0.4012&0.3244&0.4153&0.3532&{\bf 0.6417}&+ 19.3860\%\\
    \cmidrule{2-11}
    &\multicolumn{1}{|c|}{Visual Mean}&\underline{0.5088}&0.4932&0.4865&0.5148&0.3625&0.3657&0.3206&{\bf 0.6091}&+ 9.7107\%\\
\midrule
\parbox[t]{2mm}{\multirow{5}{*}{\rotatebox[origin=c]{90}{{\footnotesize text-intensive}}}} &
    \multicolumn{1}{|c|}{Code reasoning}       &0.3689&\underline{0.4085}&0.4043&0.3690&0.2923&0.2963&0.1975&{\bf 0.5376}&+ 31.6034\%\\
    &\multicolumn{1}{|c|}{Numerical calc.}       &0.3652&0.3958&0.3940&0.4241&0.3474&\underline{0.4409}&0.3423&{\bf 0.5160}&+ 17.0333\%\\
    &\multicolumn{1}{|c|}{Text translation}&\underline{0.4480}&0.3949&0.4333&0.3803&0.0931&0.2372&0.1981&{\bf 0.6196}&+ 38.3036\%\\
    &\multicolumn{1}{|c|}{OCR}        &0.4382&0.4329&0.3334&\underline{0.4455}&0.2663&0.3371&0.2912&{\bf 0.4696}&+ 5.4097\%\\
    \cmidrule{2-11}
    &\multicolumn{1}{|c|}{Textual Mean}&0.4051&\underline{0.4080}&0.3913&0.4047&0.2498&0.3279&0.2573&{\bf 0.5357}&+ 23.0875\%\\
\midrule
\multicolumn{2}{c|}{Overall Mean}
    &0.4792&0.4688&0.4593&\underline{0.4834}&0.3303&0.3549&0.3025&{\bf 0.5882}&+ 13.5327\%\\
\bottomrule
\end{tabular}%
\caption{The performance of VLLMs and humans on the GC@$1$ metric, evaluated using 5 randomly drawn images per sample/image category. The best performance achieved by an VLLM is \underline{underlined}.}
\label{table:main-gc1-sample-human}
\end{table*}

Figure~\ref{fig:correlations} presents a correlation matrix among GC@$T$ scores and the benchmarks mentioned above, where the overall GC@$T$ scores are averaged over the GC@$T$ scores of all MME categories. 
The strong correlations with the OpenCompass scores incorporating the results of multiple meticulously crafted benchmarks indicate that \textit{MMECeption} provides a comprehensive evaluation that may complement existing benchmarks. 
Further, GenCeption appears to effectively measure a VLLM's tendency to hallucinate, as demonstrated by the strong correlations with HallusionBench. 
While these observations are further emphasized by the correlations with MMStar and AI2D, the only moderate correlations with MME and SEEDBench provide more nuanced insights. 
As MME displays these moderate correlations also with the other benchmarks, it can be reasoned that it measures dimensions supplement to those measured by other benchmarks and GenCeption. 
SEEDBench on the other hand correlates strongly with other benchmarks, but only moderately with GC@$T$ scores. This indicates that SEEDBench measures aspects that are also measured by other benchmarks, but fail to be captured by GenCeption. Future research could focus on identifying these aspects to potentially incorporate them into GenCeption.

One of the key strengths of GenCeption lies in its annotation-free evaluation methodology, a concept also reflected in emerging evaluation methods such as CrossCheckGPT~ \cite{sun2024crosscheckgpt}. CrossCheckGPT ranks hallucinations by evaluating the consistency of outputs across independent MLLM models. In~\ref{tab:metric_comparison}, we analyze the correlation of CrossCheckGPT with GenCeption, MME, and HallusionBench scores. The results show strong correlations between CrossCheckGPT and both GenCeption and HallusionBench, affirming its capability to capture key evaluative metrics. Notably, CrossCheckGPT exhibits a weaker correlation with MME, which is likely because the GenCeption benchmark is developed using MME image samples, making it inherently more aligned with the MME framework.

GC@$T$ scores, as defined in Equation~\eqref{eq:gct}, are weighted by a temporal factor $t$. To examine the impact of this weighting, we conducted an ablation study where the weighting mechanism was removed, effectively transforming GC@$T$ into $s^{(T)}$. Table~\ref{tab:metric_comparison} demonstrates that $s^{(3)}$ retains a high correlation with GC@3, yet its correlation with MME diminishes compared to the weighted version, while its alignment with HallusionBench remains consistent. Furthermore, unweighted scores correlate with MME in a uniform manner across different iterations, whereas the weighted scores show a progressive increase in correlation with MME as more iterations are applied. This indicates that temporal weighting amplifies later iterations’ influence, emphasizing cumulative semantic shifts captured by MME’s iterative design. Generally, stronger correlation with MME is desirable as it validates the alignment between GenCeption’s metrics and an established benchmark, reinforcing GenCeption’s ability to assess iterative semantic coherence effectively and reliably.

Table~\ref{table:main-gc1-dalle3-imagen2} compares GC@$1$ scores using different image generators, OpenAI's DALL·E 3~\citep{ramesh2021zero} and Google DeepMind's Imagen2~\citep{deepmind2023imagegen2}. Independent of image generator used, the rankings remain unchanged, except that on visual-intensive samples only, Claude3-Opus scores equally with GPT-4V. This provides evidence that even though DALL·E 3, GPT-4o, and GPT-4V were developed and trained by OpenAI, neither of OpenAI's models has an advantage over non-OpenAI VLLMs.

Table~\ref{table:main-gc1-sample-human} shows human performance on a subset of 5 randomly drawn images per category compared to the VLLM performance on the same subset of samples. It can be observed that humans outperform all VLLMs, with especially strong differences in performance for the text-intensive categories. The worst performance, relative to humans, is achieved on the \textit{code reasoning} and \textit{text translation} categories, the former containing images of code snippets and the latter of phrases written in simplified Chinese characters.
The relatively best performance by VLLMs is achieved on the \textit{scene} and \textit{artwork} categories, which contain every-day life photos and popular artworks.
This demonstrates that there is still substantial space for performance improvement, and that compared to humans, VLLMs still lack relevant competences.
It must be noted that human performance here does not constitute an upper bound in possible scores to achieve, and that future generations of VLLMs may well outperform humans. 

\subsection{Qualitative Results}
\begin{figure*}[t!]
\centering
{\includegraphics[width=0.487\textwidth]{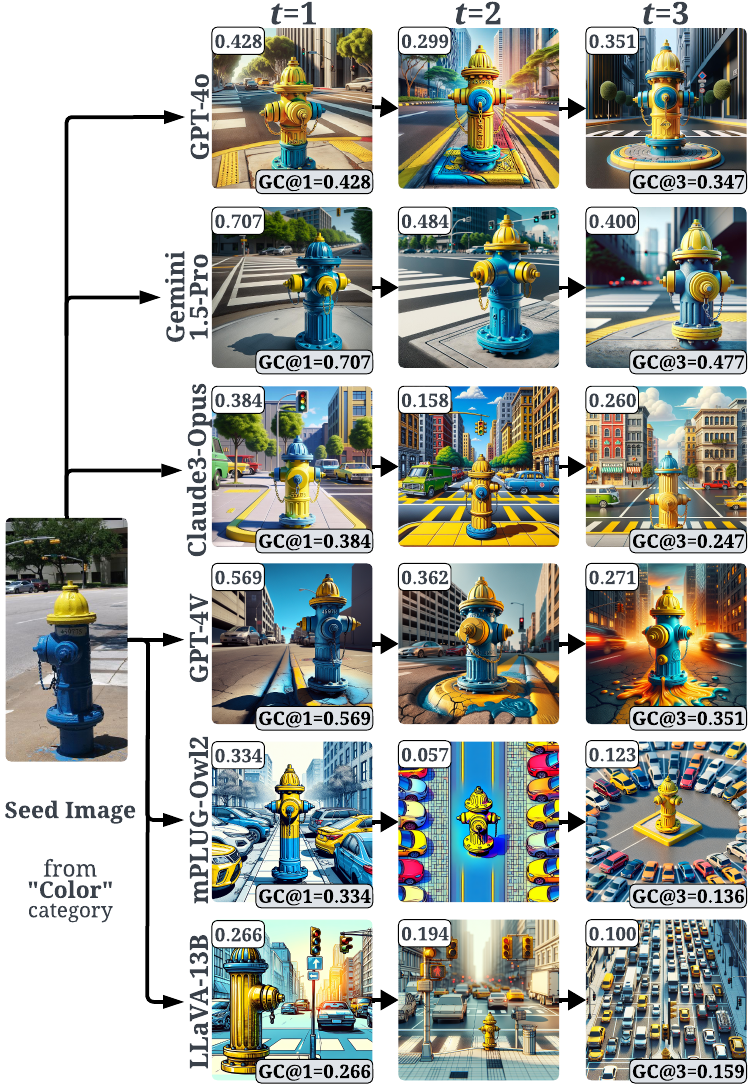}}
\hspace{6pt}
{\includegraphics[width=0.48\textwidth]{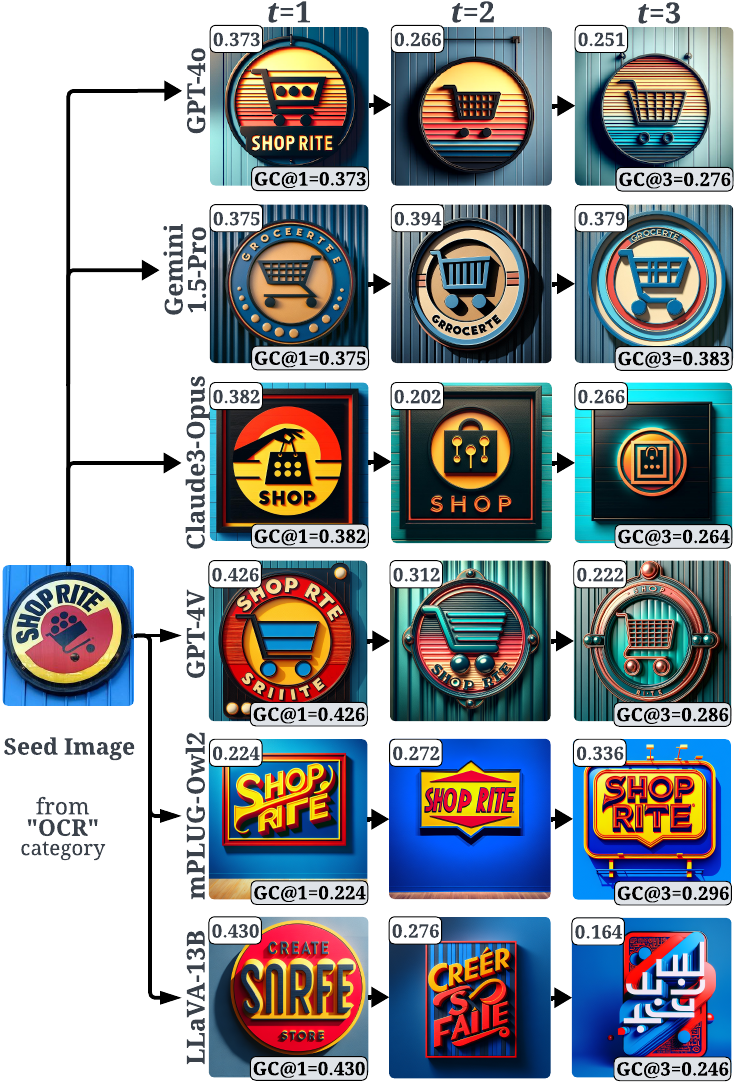}}
\caption{\label{fig:qualitative-examples}Demonstration of GenCeption evaluation procedure: the images generated over 3 GenCeption iterations for several MLLMs. The similarity $s^{(t)}$ scores (to the seed image) are shown on the top of images; GC@$1$ and GC@$3$ scores are printed on the bottom of the first and third image, respectively.}
\vspace{-10pt}
\end{figure*}

We qualitatively inspect our results by visualizing generated images together with their cosine similarity and GC@$T$ scores for two seed images across different categories, as shown in Figure~\ref{fig:qualitative-examples}. This visualization highlights the correlation between these scores and the visual characteristics of the images relative to the seed image. A key observation is that later iterations show an increased tendency to produce imagery deviating from the seed image, as indicated by lower GC@$T$ scores. This serves as an additional qualitative validation of the GenCeption method and the \textit{MMECeption} benchmark, as using VLLMs scoring higher on the \textit{MMECeption} benchmark results in generated images that preserve more information from the seed image.
For a wider range of examples across MME image categories and corresponding descriptions from each evaluated VLLM, readers are referred to \ref{sec:appendix-examples}.

\section{Discussion and Future Directions}
\label{sec:discussion}

This study validates the GenCeption method with a focus on the visual modality primarily because (1) VLLMs are the most widely used and readily available MLLMs on the market, and (2) image generation and embedding tools have reached a mature and highly commercialized stage compared to other modalities.
However, GenCeption is designed to be modality-agnostic. The same iterative procedure (i.e., describing a unimodal sample and then re-generating it from the description) can, in principle, be applied to other non-text modalities like audio and video. 
The requirement is that (1) a generation model exists for the given modality, and (2) there is a suitable encoder to quantify the similarity between the original sample and the regenerated one. 
Moreover, recent advancements have introduced multimodal LLMs capable of both generating and interpreting multiple modalities simultaneously, such as Show-o~\citep{xie2024show}, Emu3~\citep{wang2024emu3}, and JanusPro~\citep{chen2025janus}. In these cases, GenCeption could leverage the same MLLM for both description and generation tasks, serving as a particularly valuable approach for directly measuring modality consistency within such unified multimodal systems.

Future research is invited to adapt GenCeption to other non-text modalities, such as audio, video, and graphs. 
For instance, the framework can be initiated with a dataset of audio samples, and MLLMs can iteratively generate and describe the audio content. 
Similarly, for video and graph data, the process can involve generating textual descriptions of short video clips or graph structures and their recreation. 
While the core iterative process of GenCeption remains applicable, these extensions require careful exploration of modality-specific generation and embedding models.

The broad skill assessment provided by GenCeption goes along with the limitation that it is difficult to assess which skills contribute most to a high GC@$T$ score. Our analysis indicates that contemporary VLLMs perform poorly on text-intensive tasks while excelling in describing scenes and artworks. Future research could investigate this in a more fine-grained manner by creating datasets requiring specialized skills. For example, datasets could include images of complex emotions, dynamic movements, mechanical processes, or user interfaces. Additionally, combining GenCeption with specifically designed similarity metrics may offer more detailed insights into specific MLLM abilities.

\section{Conclusion}

In this paper, we introduce GenCeption to enhance the evaluation of rapidly evolving Multimodal Language Models (MLLMs). The GenCeption method attempts to address key limitations of existing MLLM benchmarks, such as costly data annotation, leading questions, the illusion of emergent abilities, and, as it allows to use newly created images without annotation, training data contamination. Further, it is expected to result in slower benchmark saturation. Being adaptable to different modalities, the GenCeption method can deliver value as a unified MLLM evaluation method that complements existing MLLM benchmarks.

Our empirical validation using the \textit{MMECeption} benchmark shows that GenCeption effectively assesses semantic coherence and consistency across modalities, aligning with established VLLM benchmarks.
By assessing humans on the \textit{MMECeption} task, we demonstrate that current VLLMs significantly lag behind human performance, particularly when working with text-intensive images.
Future work is encouraged to refine and extend this framework across a wider range of modalities, datasets, and similarity metrics.

\bibliographystyle{main} 
\bibliography{main} 

\begin{thebibliography}{36}
\expandafter\ifx\csname natexlab\endcsname\relax\def\natexlab#1{#1}\fi
\providecommand{\url}[1]{\texttt{#1}}
\providecommand{\href}[2]{#2}
\providecommand{\path}[1]{#1}
\providecommand{\DOIprefix}{doi:}
\providecommand{\ArXivprefix}{arXiv:}
\providecommand{\URLprefix}{URL: }
\providecommand{\Pubmedprefix}{pmid:}
\providecommand{\doi}[1]{\href{http://dx.doi.org/#1}{\path{#1}}}
\providecommand{\Pubmed}[1]{\href{pmid:#1}{\path{#1}}}
\providecommand{\bibinfo}[2]{#2}
\ifx\xfnm\relax \def\xfnm[#1]{\unskip,\space#1}\fi
\bibitem[{Achiam et~al.(2023)Achiam, Adler, Agarwal, Ahmad, Akkaya, Aleman, Almeida, Altenschmidt, Altman, Anadkat et~al.}]{achiam2023gpt}
\bibinfo{author}{Achiam, J.}, \bibinfo{author}{Adler, S.}, \bibinfo{author}{Agarwal, S.}, \bibinfo{author}{Ahmad, L.}, \bibinfo{author}{Akkaya, I.}, \bibinfo{author}{Aleman, F.L.}, \bibinfo{author}{Almeida, D.}, \bibinfo{author}{Altenschmidt, J.}, \bibinfo{author}{Altman, S.}, \bibinfo{author}{Anadkat, S.}, et~al., \bibinfo{year}{2023}.
\newblock \bibinfo{title}{{GPT}-4 technical report}.
\newblock \bibinfo{journal}{arXiv preprint arXiv:2303.08774} .
\bibitem[{Anthropic(2023)}]{anthropic2023claude}
\bibinfo{author}{Anthropic}, \bibinfo{year}{2023}.
\newblock \bibinfo{title}{Model card for claude}.
\newblock \URLprefix \url{https://www-cdn.anthropic.com/de8ba9b01c9ab7cbabf5c33b80b7bbc618857627/Model_Card_Claude_3.pdf}. \bibinfo{note}{accessed: [13.05.2024]}.
\bibitem[{Chen et~al.(2024)Chen, Li, Dong, Zhang, Zang, Chen, Duan, Wang, Qiao, Lin et~al.}]{chen2024we}
\bibinfo{author}{Chen, L.}, \bibinfo{author}{Li, J.}, \bibinfo{author}{Dong, X.}, \bibinfo{author}{Zhang, P.}, \bibinfo{author}{Zang, Y.}, \bibinfo{author}{Chen, Z.}, \bibinfo{author}{Duan, H.}, \bibinfo{author}{Wang, J.}, \bibinfo{author}{Qiao, Y.}, \bibinfo{author}{Lin, D.}, et~al., \bibinfo{year}{2024}.
\newblock \bibinfo{title}{Are we on the right way for evaluating large vision-language models?}
\newblock \bibinfo{journal}{arXiv preprint arXiv:2403.20330} .
\bibitem[{Chen et~al.(2025)Chen, Wu, Liu, Pan, Liu, Xie, Yu and Ruan}]{chen2025janus}
\bibinfo{author}{Chen, X.}, \bibinfo{author}{Wu, Z.}, \bibinfo{author}{Liu, X.}, \bibinfo{author}{Pan, Z.}, \bibinfo{author}{Liu, W.}, \bibinfo{author}{Xie, Z.}, \bibinfo{author}{Yu, X.}, \bibinfo{author}{Ruan, C.}, \bibinfo{year}{2025}.
\newblock \bibinfo{title}{{Janus-Pro}: Unified multimodal understanding and generation with data and model scaling}.
\newblock \bibinfo{journal}{arXiv preprint} \href{http://arxiv.org/abs/2501.17811}{{\tt arXiv:2501.17811}}.
\bibitem[{Coates et~al.(2011)Coates, Ng and Lee}]{coates2011analysis}
\bibinfo{author}{Coates, A.}, \bibinfo{author}{Ng, A.}, \bibinfo{author}{Lee, H.}, \bibinfo{year}{2011}.
\newblock \bibinfo{title}{An analysis of single-layer networks in unsupervised feature learning}, in: \bibinfo{booktitle}{Proceedings of the fourteenth international conference on artificial intelligence and statistics}, \bibinfo{organization}{JMLR Workshop and Conference Proceedings}. pp. \bibinfo{pages}{215--223}.
\bibitem[{Dai et~al.(2023)Dai, Li, Li, Tiong, Zhao, Wang, Li, Fung and Hoi}]{dai2023instructblip}
\bibinfo{author}{Dai, W.}, \bibinfo{author}{Li, J.}, \bibinfo{author}{Li, D.}, \bibinfo{author}{Tiong, A.M.H.}, \bibinfo{author}{Zhao, J.}, \bibinfo{author}{Wang, W.}, \bibinfo{author}{Li, B.}, \bibinfo{author}{Fung, P.}, \bibinfo{author}{Hoi, S.}, \bibinfo{year}{2023}.
\newblock \bibinfo{title}{Instructblip: Towards general-purpose vision-language models with instruction tuning}.
\newblock \bibinfo{journal}{arXiv preprint:2305.06500} .
\bibitem[{DeepMind(2023)}]{deepmind2023imagegen2}
\bibinfo{author}{DeepMind}, \bibinfo{year}{2023}.
\newblock \bibinfo{title}{Imagegen2}.
\newblock \bibinfo{howpublished}{\url{https://deepmind.google/technologies/imagen-2/}}.
\newblock \bibinfo{note}{Accessed: [13.05.2024]}.
\bibitem[{Deng et~al.(2009)Deng, Dong, Socher, Li, Li and Fei-Fei}]{deng2009imagenet}
\bibinfo{author}{Deng, J.}, \bibinfo{author}{Dong, W.}, \bibinfo{author}{Socher, R.}, \bibinfo{author}{Li, L.J.}, \bibinfo{author}{Li, K.}, \bibinfo{author}{Fei-Fei, L.}, \bibinfo{year}{2009}.
\newblock \bibinfo{title}{{ImageNet}: A large-scale hierarchical image database}, in: \bibinfo{booktitle}{2009 IEEE conference on computer vision and pattern recognition}, pp. \bibinfo{pages}{248--255}.
\bibitem[{Dodge et~al.(2021)Dodge, Sap, Marasovi{\'c}, Agnew, Ilharco, Groeneveld, Mitchell and Gardner}]{dodge2021documenting}
\bibinfo{author}{Dodge, J.}, \bibinfo{author}{Sap, M.}, \bibinfo{author}{Marasovi{\'c}, A.}, \bibinfo{author}{Agnew, W.}, \bibinfo{author}{Ilharco, G.}, \bibinfo{author}{Groeneveld, D.}, \bibinfo{author}{Mitchell, M.}, \bibinfo{author}{Gardner, M.}, \bibinfo{year}{2021}.
\newblock \bibinfo{title}{Documenting large webtext corpora: A case study on the colossal clean crawled corpus}.
\newblock \bibinfo{journal}{arXiv preprint arXiv:2104.08758} .
\bibitem[{Dosovitskiy et~al.(2021)Dosovitskiy, Beyer, Kolesnikov, Weissenborn, Zhai, Unterthiner, Dehghani, Minderer, Heigold, Gelly, Uszkoreit and Houlsby}]{dosovitskiy2021an}
\bibinfo{author}{Dosovitskiy, A.}, \bibinfo{author}{Beyer, L.}, \bibinfo{author}{Kolesnikov, A.}, \bibinfo{author}{Weissenborn, D.}, \bibinfo{author}{Zhai, X.}, \bibinfo{author}{Unterthiner, T.}, \bibinfo{author}{Dehghani, M.}, \bibinfo{author}{Minderer, M.}, \bibinfo{author}{Heigold, G.}, \bibinfo{author}{Gelly, S.}, \bibinfo{author}{Uszkoreit, J.}, \bibinfo{author}{Houlsby, N.}, \bibinfo{year}{2021}.
\newblock \bibinfo{title}{An image is worth 16x16 words: Transformers for image recognition at scale}, in: \bibinfo{booktitle}{International Conference on Learning Representations}.
\bibitem[{Fu et~al.(2023)Fu, Chen, Shen, Qin, Zhang, Lin, Yang, Zheng, Li, Sun et~al.}]{fu2023mme}
\bibinfo{author}{Fu, C.}, \bibinfo{author}{Chen, P.}, \bibinfo{author}{Shen, Y.}, \bibinfo{author}{Qin, Y.}, \bibinfo{author}{Zhang, M.}, \bibinfo{author}{Lin, X.}, \bibinfo{author}{Yang, J.}, \bibinfo{author}{Zheng, X.}, \bibinfo{author}{Li, K.}, \bibinfo{author}{Sun, X.}, et~al., \bibinfo{year}{2023}.
\newblock \bibinfo{title}{{MME}: A comprehensive evaluation benchmark for multimodal large language models}.
\newblock \bibinfo{journal}{arXiv preprint arXiv:2306.13394} .
\bibitem[{Heusel et~al.(2017)Heusel, Ramsauer, Unterthiner, Nessler and Hochreiter}]{heusel2017gans}
\bibinfo{author}{Heusel, M.}, \bibinfo{author}{Ramsauer, H.}, \bibinfo{author}{Unterthiner, T.}, \bibinfo{author}{Nessler, B.}, \bibinfo{author}{Hochreiter, S.}, \bibinfo{year}{2017}.
\newblock \bibinfo{title}{Gans trained by a two time-scale update rule converge to a local nash equilibrium}.
\newblock \bibinfo{journal}{Advances in neural information processing systems} \bibinfo{volume}{30}.
\bibitem[{Jiang et~al.(2023)Jiang, Chan, Chen and Wang}]{jiang2023lion}
\bibinfo{author}{Jiang, Y.}, \bibinfo{author}{Chan, C.}, \bibinfo{author}{Chen, M.}, \bibinfo{author}{Wang, W.}, \bibinfo{year}{2023}.
\newblock \bibinfo{title}{Lion: Adversarial distillation of closed-source large language model}.
\newblock \bibinfo{journal}{arXiv preprint arXiv:2305.12870} .
\bibitem[{Kembhavi et~al.(2016)Kembhavi, Salvato, Kolve, Seo, Hajishirzi and Farhadi}]{kembhavi2016diagram}
\bibinfo{author}{Kembhavi, A.}, \bibinfo{author}{Salvato, M.}, \bibinfo{author}{Kolve, E.}, \bibinfo{author}{Seo, M.}, \bibinfo{author}{Hajishirzi, H.}, \bibinfo{author}{Farhadi, A.}, \bibinfo{year}{2016}.
\newblock \bibinfo{title}{A diagram is worth a dozen images}, in: \bibinfo{booktitle}{Computer Vision--ECCV 2016: 14th European Conference, Amsterdam, The Netherlands, October 11--14, 2016, Proceedings, Part IV 14}, \bibinfo{organization}{Springer}. pp. \bibinfo{pages}{235--251}.
\bibitem[{Kiela et~al.(2021)Kiela, Bartolo, Nie, Kaushik, Geiger, Wu, Vidgen, Prasad, Singh, Ringshia et~al.}]{kiela2021dynabench}
\bibinfo{author}{Kiela, D.}, \bibinfo{author}{Bartolo, M.}, \bibinfo{author}{Nie, Y.}, \bibinfo{author}{Kaushik, D.}, \bibinfo{author}{Geiger, A.}, \bibinfo{author}{Wu, Z.}, \bibinfo{author}{Vidgen, B.}, \bibinfo{author}{Prasad, G.}, \bibinfo{author}{Singh, A.}, \bibinfo{author}{Ringshia, P.}, et~al., \bibinfo{year}{2021}.
\newblock \bibinfo{title}{Dynabench: Rethinking benchmarking in nlp}, in: \bibinfo{booktitle}{Proceedings of the 2021 Conference of the North American Chapter of the Association for Computational Linguistics: Human Language Technologies}, pp. \bibinfo{pages}{4110--4124}.
\bibitem[{Krizhevsky et~al.(2009)Krizhevsky, Hinton et~al.}]{krizhevsky2009learning}
\bibinfo{author}{Krizhevsky, A.}, \bibinfo{author}{Hinton, G.}, et~al., \bibinfo{year}{2009}.
\newblock \bibinfo{title}{Learning multiple layers of features from tiny images}.
\newblock \bibinfo{type}{Technical Report}. Massachusetts Institute of Technology and New York University.
\bibitem[{Lee et~al.(2024)Lee, Park and Kang}]{lee-etal-2024-fleur}
\bibinfo{author}{Lee, Y.}, \bibinfo{author}{Park, I.}, \bibinfo{author}{Kang, M.}, \bibinfo{year}{2024}.
\newblock \bibinfo{title}{{FLEUR}: An explainable reference-free evaluation metric for image captioning using a large multimodal model}, in: \bibinfo{editor}{Ku, L.W.}, \bibinfo{editor}{Martins, A.}, \bibinfo{editor}{Srikumar, V.} (Eds.), \bibinfo{booktitle}{Proceedings of the 62nd Annual Meeting of the Association for Computational Linguistics (Volume 1: Long Papers)}, \bibinfo{publisher}{Association for Computational Linguistics}, \bibinfo{address}{Bangkok, Thailand}. pp. \bibinfo{pages}{3732--3746}.
\newblock \URLprefix \url{https://aclanthology.org/2024.acl-long.205}, \DOIprefix\doi{10.18653/v1/2024.acl-long.205}.
\bibitem[{Li et~al.(2023a)Li, Wang, Wang, Ge, Ge and Shan}]{li2023seed}
\bibinfo{author}{Li, B.}, \bibinfo{author}{Wang, R.}, \bibinfo{author}{Wang, G.}, \bibinfo{author}{Ge, Y.}, \bibinfo{author}{Ge, Y.}, \bibinfo{author}{Shan, Y.}, \bibinfo{year}{2023}a.
\newblock \bibinfo{title}{Seed-bench: Benchmarking multimodal llms with generative comprehension}.
\newblock \bibinfo{journal}{arXiv preprint arXiv:2307.16125} .
\bibitem[{Li et~al.(2023b)Li, Patel and Du}]{li2023prd}
\bibinfo{author}{Li, R.}, \bibinfo{author}{Patel, T.}, \bibinfo{author}{Du, X.}, \bibinfo{year}{2023}b.
\newblock \bibinfo{title}{Prd: Peer rank and discussion improve large language model based evaluations}.
\newblock \bibinfo{journal}{arXiv preprint arXiv:2307.02762} .
\bibitem[{Li et~al.(2023c)Li, Du, Zhou, Wang, Zhao and Wen}]{li2023evaluating}
\bibinfo{author}{Li, Y.}, \bibinfo{author}{Du, Y.}, \bibinfo{author}{Zhou, K.}, \bibinfo{author}{Wang, J.}, \bibinfo{author}{Zhao, W.X.}, \bibinfo{author}{Wen, J.R.}, \bibinfo{year}{2023}c.
\newblock \bibinfo{title}{Evaluating object hallucination in large vision-language models}.
\newblock \bibinfo{journal}{arXiv preprint:2305.10355} .
\bibitem[{Liu et~al.(2023a)Liu, Guan, Li, Chen, Yacoob, Manocha and Zhou}]{liu2023hallusionbench}
\bibinfo{author}{Liu, F.}, \bibinfo{author}{Guan, T.}, \bibinfo{author}{Li, Z.}, \bibinfo{author}{Chen, L.}, \bibinfo{author}{Yacoob, Y.}, \bibinfo{author}{Manocha, D.}, \bibinfo{author}{Zhou, T.}, \bibinfo{year}{2023}a.
\newblock \bibinfo{title}{Hallusionbench: You see what you think? or you think what you see? an image-context reasoning benchmark challenging for gpt-4v (ision), llava-1.5, and other multi-modality models}.
\newblock \bibinfo{journal}{arXiv preprint arXiv:2310.14566} .
\bibitem[{Liu et~al.(2023b)Liu, Li, Li and Lee}]{liu2023improved}
\bibinfo{author}{Liu, H.}, \bibinfo{author}{Li, C.}, \bibinfo{author}{Li, Y.}, \bibinfo{author}{Lee, Y.J.}, \bibinfo{year}{2023}b.
\newblock \bibinfo{title}{Improved baselines with visual instruction tuning}, in: \bibinfo{booktitle}{NeurIPS 2023 Workshop on Instruction Tuning and Instruction Following}.
\bibitem[{Loya et~al.(2023)Loya, Sinha and Futrell}]{loya2023exploring}
\bibinfo{author}{Loya, M.}, \bibinfo{author}{Sinha, D.A.}, \bibinfo{author}{Futrell, R.}, \bibinfo{year}{2023}.
\newblock \bibinfo{title}{Exploring the sensitivity of llms' decision-making capabilities: Insights from prompt variation and hyperparameters}.
\newblock \bibinfo{journal}{arXiv preprint arXiv:2312.17476} .
\bibitem[{OpenAI(2024)}]{openai2024gpt4o}
\bibinfo{author}{OpenAI}, \bibinfo{year}{2024}.
\newblock \bibinfo{title}{Hello gpt-4o}.
\newblock \bibinfo{journal}{OpenAI Technical Report} \bibinfo{note}{Available at \url{https://openai.com/index/hello-gpt-4o/}}.
\bibitem[{OpenCompass(2023)}]{2023opencompass}
\bibinfo{author}{OpenCompass}, \bibinfo{year}{2023}.
\newblock \bibinfo{title}{{OpenCompass}: A universal evaluation platform for foundation models}.
\newblock \bibinfo{howpublished}{\url{https://github.com/open-compass/opencompass}}.
\bibitem[{Ramesh et~al.(2021)Ramesh, Pavlov, Goh, Gray, Voss, Radford, Chen and Sutskever}]{ramesh2021zero}
\bibinfo{author}{Ramesh, A.}, \bibinfo{author}{Pavlov, M.}, \bibinfo{author}{Goh, G.}, \bibinfo{author}{Gray, S.}, \bibinfo{author}{Voss, C.}, \bibinfo{author}{Radford, A.}, \bibinfo{author}{Chen, M.}, \bibinfo{author}{Sutskever, I.}, \bibinfo{year}{2021}.
\newblock \bibinfo{title}{Zero-shot text-to-image generation}, in: \bibinfo{booktitle}{International Conference on Machine Learning}, \bibinfo{organization}{PMLR}. pp. \bibinfo{pages}{8821--8831}.
\bibitem[{Reid et~al.(2024)Reid, Savinov, Teplyashin, Lepikhin, Lillicrap, Alayrac, Soricut, Lazaridou, Firat, Schrittwieser et~al.}]{reid2024gemini}
\bibinfo{author}{Reid, M.}, \bibinfo{author}{Savinov, N.}, \bibinfo{author}{Teplyashin, D.}, \bibinfo{author}{Lepikhin, D.}, \bibinfo{author}{Lillicrap, T.}, \bibinfo{author}{Alayrac, J.b.}, \bibinfo{author}{Soricut, R.}, \bibinfo{author}{Lazaridou, A.}, \bibinfo{author}{Firat, O.}, \bibinfo{author}{Schrittwieser, J.}, et~al., \bibinfo{year}{2024}.
\newblock \bibinfo{title}{Gemini 1.5: Unlocking multimodal understanding across millions of tokens of context}.
\newblock \bibinfo{journal}{arXiv preprint arXiv:2403.05530} .
\bibitem[{Schaeffer et~al.(2023)Schaeffer, Miranda and Koyejo}]{schaeffer2023emergent}
\bibinfo{author}{Schaeffer, R.}, \bibinfo{author}{Miranda, B.}, \bibinfo{author}{Koyejo, S.}, \bibinfo{year}{2023}.
\newblock \bibinfo{title}{Are emergent abilities of large language models a mirage?}
\newblock \bibinfo{journal}{arXiv preprint arXiv:2304.15004} .
\bibitem[{Sun et~al.(2024)Sun, Manakul, Liusie, Pipatanakul, Zhang, Woodland and Gales}]{sun2024crosscheckgpt}
\bibinfo{author}{Sun, G.}, \bibinfo{author}{Manakul, P.}, \bibinfo{author}{Liusie, A.}, \bibinfo{author}{Pipatanakul, K.}, \bibinfo{author}{Zhang, C.}, \bibinfo{author}{Woodland, P.}, \bibinfo{author}{Gales, M.}, \bibinfo{year}{2024}.
\newblock \bibinfo{title}{Crosscheckgpt: Universal hallucination ranking for multimodal foundation models}.
\newblock \bibinfo{journal}{arXiv preprint arXiv:2405.13684} .
\bibitem[{Wang et~al.(2023)Wang, Chen, Chen, Wu, Zhu, Zeng, Luo, Lu, Zhou, Qiao et~al.}]{wang2023visionllm}
\bibinfo{author}{Wang, W.}, \bibinfo{author}{Chen, Z.}, \bibinfo{author}{Chen, X.}, \bibinfo{author}{Wu, J.}, \bibinfo{author}{Zhu, X.}, \bibinfo{author}{Zeng, G.}, \bibinfo{author}{Luo, P.}, \bibinfo{author}{Lu, T.}, \bibinfo{author}{Zhou, J.}, \bibinfo{author}{Qiao, Y.}, et~al., \bibinfo{year}{2023}.
\newblock \bibinfo{title}{Visionllm: Large language model is also an open-ended decoder for vision-centric tasks}.
\newblock \bibinfo{journal}{arXiv preprint:2305.11175} .
\bibitem[{Wang et~al.(2024)Wang, Zhang, Luo, Sun, Cui, Wang, Zhang, Wang, Li, Yu et~al.}]{wang2024emu3}
\bibinfo{author}{Wang, X.}, \bibinfo{author}{Zhang, X.}, \bibinfo{author}{Luo, Z.}, \bibinfo{author}{Sun, Q.}, \bibinfo{author}{Cui, Y.}, \bibinfo{author}{Wang, J.}, \bibinfo{author}{Zhang, F.}, \bibinfo{author}{Wang, Y.}, \bibinfo{author}{Li, Z.}, \bibinfo{author}{Yu, Q.}, et~al., \bibinfo{year}{2024}.
\newblock \bibinfo{title}{Emu3: Next-token prediction is all you need}.
\newblock \bibinfo{journal}{arXiv preprint} \href{http://arxiv.org/abs/2409.18869}{{\tt arXiv:2409.18869}}.
\bibitem[{Xie et~al.(2024)Xie, Mao, Bai, Zhang, Wang, Lin, Gu, Chen, Yang and Shou}]{xie2024show}
\bibinfo{author}{Xie, J.}, \bibinfo{author}{Mao, W.}, \bibinfo{author}{Bai, Z.}, \bibinfo{author}{Zhang, D.J.}, \bibinfo{author}{Wang, W.}, \bibinfo{author}{Lin, K.Q.}, \bibinfo{author}{Gu, Y.}, \bibinfo{author}{Chen, Z.}, \bibinfo{author}{Yang, Z.}, \bibinfo{author}{Shou, M.Z.}, \bibinfo{year}{2024}.
\newblock \bibinfo{title}{Show-o: One single transformer to unify multimodal understanding and generation}.
\newblock \bibinfo{journal}{arXiv preprint} \href{http://arxiv.org/abs/2408.12528}{{\tt arXiv:2408.12528}}.
\bibitem[{Xu et~al.(2022)Xu, Shen and Huang}]{xu2022multiinstruct}
\bibinfo{author}{Xu, Z.}, \bibinfo{author}{Shen, Y.}, \bibinfo{author}{Huang, L.}, \bibinfo{year}{2022}.
\newblock \bibinfo{title}{Multiinstruct: Improving multi-modal zero-shot learning via instruction tuning}.
\newblock \bibinfo{journal}{arXiv preprint:2212.10773} .
\bibitem[{Yang et~al.(2023)Yang, Chiang, Zheng, Gonzalez and Stoica}]{yang2023rethinking}
\bibinfo{author}{Yang, S.}, \bibinfo{author}{Chiang, W.L.}, \bibinfo{author}{Zheng, L.}, \bibinfo{author}{Gonzalez, J.E.}, \bibinfo{author}{Stoica, I.}, \bibinfo{year}{2023}.
\newblock \bibinfo{title}{Rethinking benchmark and contamination for language models with rephrased samples}.
\newblock \bibinfo{journal}{arXiv preprint arXiv:2311.04850} .
\bibitem[{Ye et~al.(2023)Ye, Xu, Ye, Yan, Liu, Qian, Zhang, Huang and Zhou}]{ye2023mplug}
\bibinfo{author}{Ye, Q.}, \bibinfo{author}{Xu, H.}, \bibinfo{author}{Ye, J.}, \bibinfo{author}{Yan, M.}, \bibinfo{author}{Liu, H.}, \bibinfo{author}{Qian, Q.}, \bibinfo{author}{Zhang, J.}, \bibinfo{author}{Huang, F.}, \bibinfo{author}{Zhou, J.}, \bibinfo{year}{2023}.
\newblock \bibinfo{title}{mplug-owl2: Revolutionizing multi-modal large language model with modality collaboration}.
\newblock \bibinfo{journal}{arXiv preprint:2311.04257} .
\bibitem[{Zhao et~al.(2023)Zhao, Pang, Du, Yang, Li, Cheung and Lin}]{zhao2023evaluating}
\bibinfo{author}{Zhao, Y.}, \bibinfo{author}{Pang, T.}, \bibinfo{author}{Du, C.}, \bibinfo{author}{Yang, X.}, \bibinfo{author}{Li, C.}, \bibinfo{author}{Cheung, N.M.}, \bibinfo{author}{Lin, M.}, \bibinfo{year}{2023}.
\newblock \bibinfo{title}{On evaluating adversarial robustness of large vision-language models}.
\newblock \bibinfo{journal}{arXiv preprint:2305.16934} .

\end{thebibliography}

\appendix

\section{GenCeption Demonstration}
\label{sec:appendix-examples}
\begin{figure}[h]
    \centering
    \centering
    \includegraphics[width=.5\linewidth]{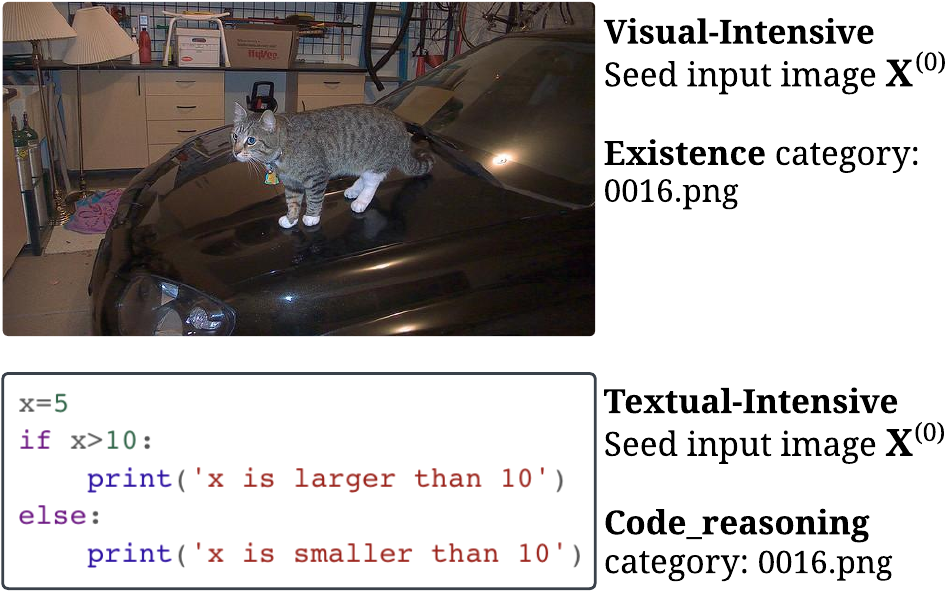}
    \caption{Example seed images from the visually (Figure~\ref{fig:appendix-visual}) and textually (Figure~\ref{fig:appendix-textual}) intensive groups, along with their associated metadata.}
    \label{fig:seed-images}
\end{figure}

\begin{figure*}
  \vspace*{-2cm} 
  \hspace{-13pt}
  \includegraphics[width=16.6cm]{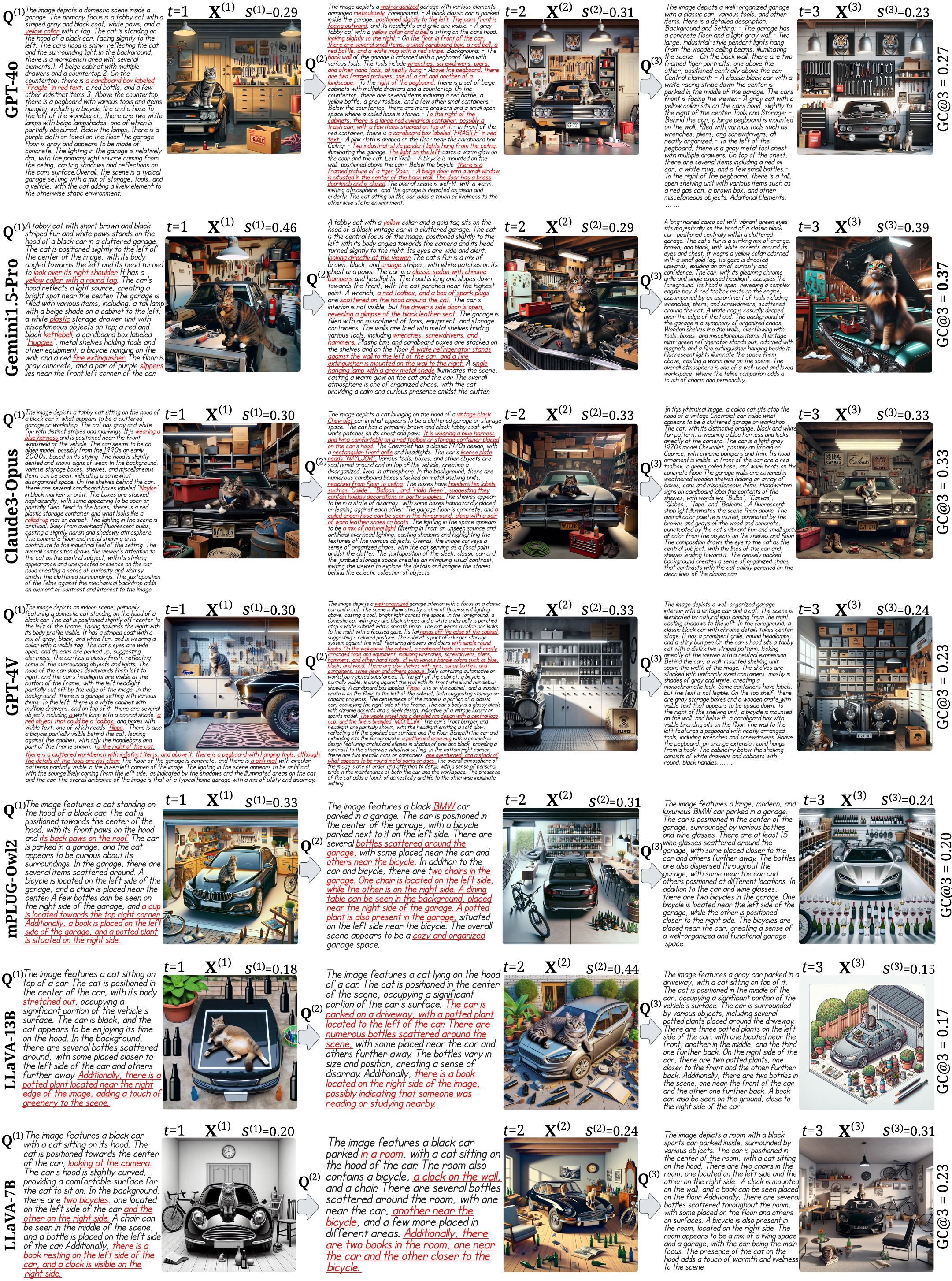}
  \vspace{-10pt}
  \caption{Illustration of a 3-iteration GenCeption procedure run on a {\bf visual-intensive} image (from ``existence'' category) to evaluate 7 VLLMs. Each iteration \(t\) shows the generated image \(\mathbf{X}^{(t)}\), the description \(\mathbf{Q}^{(t)}\) of the preceding image \(\mathbf{X}^{(t-1)}\), and the similarity score \(s^{(t)}\) relative to \(\mathbf{X}^{(0)}\). The GC@$3$ metric for each VLLM is also presented. Hallucinated elements within descriptions \(\mathbf{Q}^{(1)}\) and \(\mathbf{Q}^{(2)}\) as compared to the seed image are indicated with \color{red}{\underline{red underlining}}.}
  \label{fig:appendix-visual}
\end{figure*}

\begin{figure*}
  \vspace*{-2cm} 
  \hspace{-13pt}
  \includegraphics[width=16.6cm]{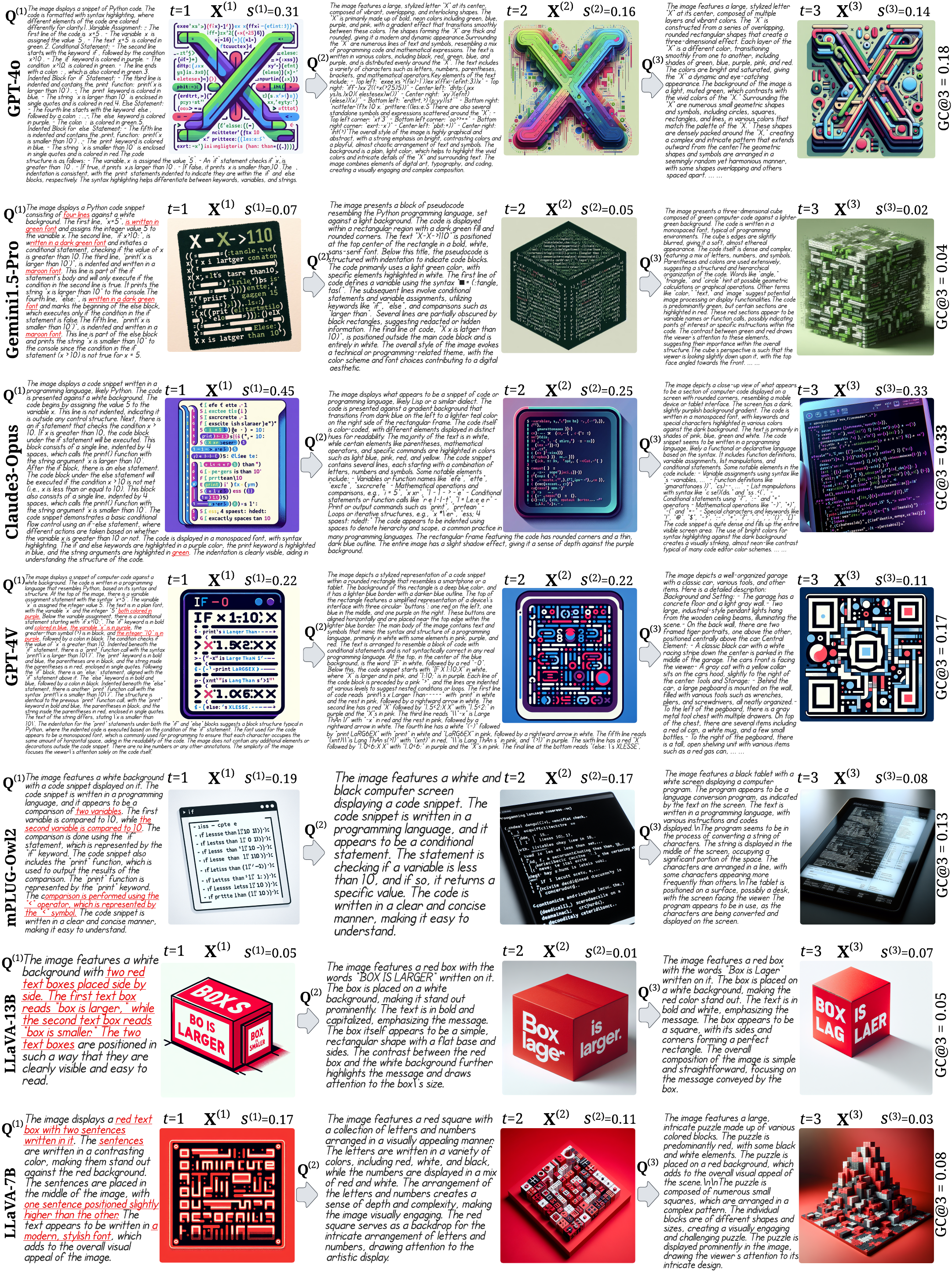}
  \vspace{-10pt}
  \caption{Illustration of a 3-iteration GenCeption procedure run on a {\bf textual-intensive} image (from ``code reasoning'' category) to evaluate 7 VLLMs. Each iteration \(t\) shows the generated image \(\mathbf{X}^{(t)}\), the description \(\mathbf{Q}^{(t)}\) of the preceding image \(\mathbf{X}^{(t-1)}\), and the similarity score \(s^{(t)}\) relative to \(\mathbf{X}^{(0)}\). The GC@$3$ metric for each VLLM is also presented. Hallucinated elements within descriptions \(\mathbf{Q}^{(1)}\) as compared to the seed image are indicated with \color{red}{\underline{red underlining}}.}
  \label{fig:appendix-textual}
\end{figure*}

To provide a comprehensive, intuitive and qualitative understanding of the GenCeption procedure and GC@\({T}\) metric, we illustrate the input, output, intermediate artifacts, similarity scores, and GC@\({T}\) values throughout the GenCeption process. 
Examples from the visual-intensive and textual-intensive groups are showcased in Figures~\ref{fig:appendix-visual} and \ref{fig:appendix-textual}, respectively. 
The corresponding seed images and their metadata are presented in Figure~\ref{fig:seed-images}.

\section{Dataset and Reproducibility}

In Sections~\ref{sec:introduction}, \ref{sec:procedure}, \ref{sec:metric} and \ref{sec:experiments} of the main paper, we cite the creators of all artifacts used. Detailed citations can be found in references. 
The MME dataset is not directly downloadable, and is released for research purposes only upon a request from authors to gain access to it. It does not contain any personally identifying information, as the questions regard visual aspects of the images.
We followed the guidelines provided by the authors and respected the intended terms of use. 
The specific licenses and terms for the use and distribution of publicly available artifacts can be found in the corresponding original papers or GitHub repositories, as cited. 
As per this research work and aligning with the MME copyrights, we are not releasing this asset. 
Regarding the created artifacts, we introduce a new metric called GC@$T$, and detail its creation and intended use in Section~\ref{sec:metric} of the main paper. 
Our study exclusively utilizes images from the MME dataset, omitting textual QA annotations, and generates textual data in the form of English descriptions as part of our methodology.
Given the nature of our research centered on quantifying the inter-modality coherence and consistency, we do not apply any data splits. Due to limitations in terms of computational resources, the metrics reported in Table~\ref{table:main-visual-text-intensive} are from a single run.

In our study, we adopt several state-of-the-art models to facilitate our experiments, including Gemini1.5-Pro, GPT-4o, GPT-4V, Claude3-Opus, LLaVa-13B, LLaVa-7B, and mPLUG-Ow12 for text description generation, ViT for image embedding, and DALL·E 3 and Imagen2 for image generation, adhering to default parameter settings as outlined in their original specifications. As we only evaluated existing models but did not train new models, no hyperparamter tuning was applicaple.
The text descriptions generated by GPT-4V/4o, Claude3 and Gemini1.5 are obtained through API calls, while experiments involving the other models are conducted on A100 GPUs, totaling approximately 96 GPU hours. 
Image generation was also performed via a call to OpenAI's DALL-E 3 API. 
To compute the GC@$T$ metric, we employ the cosine similarity metric from the Scikit-learn library (Version 1.4.0).

\section{Human Annotators}

As described in Section~\ref{sec:experiments}, we benchmarked the performance of humans at the GC@$T$ task. The 5 human annotators were recruited from the authors' social circle and were being made aware that they contributed to a research project, with the specific goal of the research project only being disclosed after participation. They were provided with the same instruction prompt as the MLLMs, and given 14 weeks to complete the task. This task does not involve sharing any personal information and the images were carefully evaluated to not be offensive in any way. The participants gave consent that their annotations could be used for scientific research and included in research papers. 

\section{AI Assistants}

AI assistants like GitHub Copilot, ChatGPT, and Perplexity were used to support writing the necessary codebase and find efficient ways to express complex concepts. AI assistant suggestions were always carefully evaluated for correctness and only used after human revision.

\end{document}